\documentclass[a4paper,fleqn]{cas-sc}

\usepackage[numbers]{natbib}
\usepackage{amsmath,amsfonts}
\usepackage{amsthm}
\usepackage{algorithmic}
\usepackage{algorithm}
\usepackage{subcaption}
\usepackage{float}
\def\tsc#1{\csdef{#1}{\textsc{\lowercase{#1}}\xspace}}
\tsc{WGM}
\tsc{QE}
\tsc{EP}
\tsc{PMS}
\tsc{BEC}
\tsc{DE}
\begin{document}
\let\WriteBookmarks\relax
%\def\floatpagepagefraction{1}
%\def\textpagefraction{.001}
% Replace those lines with:
\renewcommand{\topfraction}{0.95}
\renewcommand{\bottomfraction}{0.95}
\renewcommand{\textfraction}{0.05}
\renewcommand{\floatpagefraction}{0.70}  % Must NEVER be 1 or 1.0!
\setcounter{totalnumber}{6}

\shorttitle{Generalized Sparse Fuzzy Clustering}
\shortauthors{G. Stamatelis and K. Koutroumbas}
%\begin{frontmatter}

\title [mode = title]{A Generalized Sparse Fuzzy Clustering Framework with Automated Cluster Cardinality Estimation Ability}

%\tnotetext[2]{The second title footnote which is a longer text matter
%   to fill through the whole text width and overflow into
%   another line in the footnotes area of the first page.}

\author[1]{George Stamatelis}
\cormark[1]
\ead{georgestamat@di.uoa.gr}
%\ead[url]{www.jkkrishnan.in}

\credit{Methodology, Software, Writting}

\affiliation[1]{organization={Department of Informatics and Telecommunications, National and Kapodistrian University of Athens},
                addressline={Athens 16122}, 
                city={Athens},
%               citysep={}, % Uncomment if no comma needed between city and postcode
                %postcode={695013}, 
                state={Attica},
                country={Greece}}

\author[2]{Konstantinos  Koutroumbas}
\affiliation[2]{
organization={
Institute for Astronomy Astrophysics Space Applications and Remote Sensing, National Observatory of Athens},
addressline={Vas. Pavlou \& I. Metaxa, 15236},
city={Penteli},
state={Attica},
country={Greece}
}
\ead{koutroum@noa.gr}
\credit{Conceptualization, Writing, Supervision}

\cortext[cor1]{Corresponding author}

\begin{abstract}
Fuzzy clustering provides a robust framework for handling ambiguous and overlapping data boundaries, yet its traditional formulation generates entirely dense membership assignments that limit noise resilience and prevent estimating   the true number  of clusters, when it is unknown. To address these limitations, we introduce in this paper  the Generalized Sparse Fuzzy Clustering (GSFC) framework, which incorporates a continuous sparsity penalty into the fuzzy objective function. Because closed-form solutions are precluded by the simultaneous presence of fuzzifier and sparsity exponents, we propose an efficient primal-dual bisection search coupled with specialized analytic pathways to rigorously enforce the partition constraint while organically driving trace memberships to exact zero. This mathematically enforced sparsity empowers the algorithm to autonomously estimate the true number of clusters by dynamically pruning redundant clusters,  provided that it is initialized by a larger than the true number of clusters.   Furthermore, we integrate a virtual noise sink to insulate outliers from the rest of the data points. The usage of the proposed framework is also extended to line segment detection.  Extensive empirical evaluations across synthetic toy datasets, high-noise geometric stress tests, and real-world  benchmarks (UCI wine dataset and hyperspectral images) demonstrate that GSFC effectively exceeds conventional fuzzy benchmarks while autonomously resolving cluster cardinality. Additionally, it  significantly outperforms existing sparse possibilistic and probabilistic soft-clustering alternatives in line detection tasks. 
\end{abstract}

%\begin{graphicalabstract}
%\includegraphics{figs/cas-grabs.pdf}
%\end{graphicalabstract}

\begin{highlights}
    \item We propose the Generalized Sparse Fuzzy Clustering (GSFC) framework, introducing an additional term into the fuzzy objective, that promotes sparsity on the fuzzy membership values.
    \item A novel primal-dual optimization algorithm combining bisection search with closed-form analytic pathways  
  is developed to solve the resulting Cost Function Optimization (CFO) problem .
  \item Due to the enforced sparsity on membership values redundant clusters are pruned, implying that the algorithm can reliably infer the true number of clusters from a safe over-initialization. Henceforth, prior knowledge of the exact number of clusters is not required. 
  \item Two specific algorithmic instances for the GSFC framework, which are suitable for the identification of compact and elongated clusters, respectively, are discussed extensively. 
    \item Comprehensive evaluations across synthetic data and multiple real-world datasets, for both the compact and the elongated cluster case, showcase that the introduced sparsity penalty results in superior performance over other relevant clustering approaches.

\end{highlights}


\maketitle

\section{Introduction}
The primary focus of this paper is the introduction of sparsity on membership assignments in Cost Function Optimization (CFO) clustering algorithms. CFO algorithms have two general categories: 
(a) the hard CFO algorithms where each data point belongs exclusively to a single cluster (this branch includes the probabilistic set up~\cite{probabilistic_general}) and (b) the soft CFO algorithms where the concept of membership of a point to a cluster is relaxed (this branch includes the fuzzy~\cite{fuzzyGeneral1,fuzzyGeneral2} and possibilistic algorithms~\cite{possibilistic_general_1,possibilistic_general_2}). It has been shown that imposing sparsity in  possibilistc and probabilistic CFO algorithms greatly improves their performance~\cite{K18,S_and_K21}.
%Soft clustering generally falls into three main paradigms: probabilistic (mixture models)~\cite{probabilistic_general}, possibilistic~\cite{possibilistic_general_1,possibilistic_general_2}, and fuzzy~\cite{fuzzyGeneral1,fuzzyGeneral2}, with sparse-augmented objectives having shown to greatly improve the  performance of algorithms in the first two categories, e.g. \cite{K18,S_and_K21}.

Fuzzy clustering distinguishes itself from rigid hard-partitioning methods by allowing data points to belong to multiple clusters simultaneously up to a degree of membership, a strategy that naturally accommodates data ambiguity and overlapping boundaries. However, this mathematical flexibility inherently produces entirely dense membership assignments, meaning every data point retains a non-zero connection to every cluster. This dense modeling risks  artificially blurring boundaries between neighboring groups, heavily increasing sensitivity to noise and outliers (which persistently drag cluster representatives away from their true positions), and prevents the autonomous discovery of the dataset's true clustering structure. Because points are forced to distribute their membership across all available clusters, there is no way to identify and prune redundant clusters.
For instance, consider a line segment detection application featuring an image with 50  line segments. While a single pixel might belong to multiple intersecting line segments—necessitating soft assignments—physical reality dictates that it will intersect with only a small fraction of them, not all 50. The potential for such unrealistic representations highlights the need for a framework that can naturally drive irrelevant, trace memberships to exact zero. As will be experimentally showcased, enforcing sparsity in the data membership vectors not only improves clustering quality but  also can identify and prune redundant clusters, effectively learning the number of true clusters in the dataset.
\subsection{Literature Review}
While there have been numerous attempts at introducing sparsity into fuzzy clustering, most  works  focus on different sparsity models. For instance, Zhang et al \cite{lui2025_fuzzy_with_sparse_noise}, introduces sparsity into the general fuzzy clustering framework, focusing on image segmentation tasks. It assumes that each data point $\mathbf{x}_i$ is a noisy observation, and it learns the "classic"  (i.e., non-sparse) fuzzy membership weight $u_{i,j}$, the prototypes $\boldsymbol{\theta}_j$ and the sparse noise components $\mathbf{e}_i$. Hence, the typical issues associated with fuzzy clustering discussed above  remain.    Furthermore, its noise suppression is limited by its strict reliance on a manually tuned deviation hyperparameter and the exact number of clusters needs to be specified a priori.

Similarly, Cheng et al \cite{Sparse_Fuzzy_with_PCA}, introduces a Sparsity Fuzzy C-Means framework with Principal Component Analysis (PCA) embeddings that performs dimensionality reduction and fuzzy membership learning simultaneously. To mitigate the impact of noise in high-dimensional spaces, the authors introduce a binary weighting factor to adaptively filter out outliers (hard thresholding).  However, this  approach presents significant practical limitations, as it strictly requires the user to pre-define the exact number of normal (i.e., non-outlier) data points ($k$) present in the dataset. The authors explicitly acknowledge that overestimating $k$ allows outliers to contaminate the clusters, while underestimating $k$ permanently discards valid data points. Furthermore, the proposed method requires a  fixed, predetermined number of clusters. Consequently, sparsity in this model acts merely as a rigid, quota-based outlier rejection tool.

A closely related strategy is adopted by Zhang et al. \cite{Sparce_Fuzzy_SPL}, who propose the so-called Row Sparse Discriminative Embedding Projection (RFCDE) framework. Like \cite{Sparse_Fuzzy_with_PCA}, RFCDE learns a low-dimensional subspace while utilizing an adaptive binary sample contribution vector to filter out outliers. Although RFCDE successfully induces row-sparse memberships via entropy regularization and mitigates prototype bias through joint projection learning, it inherits the exact same fundamental vulnerability regarding the hyperparameter $k$. Because the algorithm relies on ranking sample contributions to strictly isolate $k$ normal samples, an inaccurate estimate of $k$ will inevitably exclude valid, informative data or absorb outliers into the clusters. Furthermore, it also requires the total number of clusters to be known in advance.

Another interesting direction, given in  \cite{Sparse_Fuzzy_Lasso},  builds directly upon the foundational Membership Affinity Lasso (MalFCM) framework \cite{Lasso_Fuzzy}, which attempts to preserve local data geometry by applying a Lasso penalty that forces neighboring points to share similar fuzzy assignments. To address the entirely dense memberships generated within the original MalFCM framework, the extended Sparse Membership Affinity Lasso (SMalFCM) introduces an explicit $L_0$-norm constraint to limit the maximum number of clusters a single sample can belong to. However, this approach enforces sparsity through a rigid method (a user-defined hyperparameter $\tilde{s}$) that hard-truncates memberships, rather than achieving it organically through a continuous mathematical penalty term, as is the case with  our proposed  framework. Furthermore, both SMalFCM and its predecessor rely entirely on pre-computed nearest-neighbor graphs to dictate membership affinity, making their performance highly fragile if this initial geometric prior is not accurate, a common occurrence in high-dimensional or noisy datasets. Finally, these approaches require the exact number of clusters to be strictly defined in advance and practically restrict the fuzzifier to a value of $q=2$ to maintain closed-form ADMM updates.

In \cite{fuzzy_teniary_tree} the focus is on large high-dimensional data applications, proposing the Robust Jointly Sparse Fast Fuzzy Clustering (RSFCAG) algorithm to address  computational bottlenecks. To achieve linear time complexity, RSFCAG approximates the dataset using a small subset of anchor points—generated via a fast ternary-tree hierarchical approach—and performs fuzzy clustering within a low-dimensional space obtained through joint Locality Preserving Projection (LPP). The algorithm introduces sparsity strictly for denoising and graph construction. However, approximating the entire dataset with a sparse set of anchors inevitably results in information loss, which, as the authors note, can degrade clustering performance on datasets that require fine-grained topological mapping. Furthermore, the framework strictly requires the exact number of clusters to be pre-specified and relies on extensive hyperparameter tuning. Critically, while the projection matrices and anchor connections are sparse, the actual fuzzy membership updates do not enforce sparsity, resulting in dense assignments. %Consequently, it lacks a continuous mathematical mechanism—like the one proposed in our GSFC framework—to organically drive ambiguous memberships to exact zeros, naturally shape fuzzy boundaries, and autonomously discover the true cluster cardinality.  

Reference \cite{Sparse_Fuzzy_Image_Semg} integrates conventional fuzzy methods with Sparse Subspace Clustering (SSC) to form a unified Fuzzy Sparse Subspace Clustering (FSSC) framework. By embedding the self-representation coefficients of SSC into the fuzzy objective function, FSSC leverages global data correlations to mitigate the effects of complex backgrounds and intensity inhomogeneity in infrared images. However, the construction of the self-representation matrix introduces severe computational bottlenecks, forcing the algorithm to rely on superpixel preprocessing to reduce the input size. Furthermore, the required ADMM optimization mathematically restricts the fuzzifier to $q=2$ and assumes a rigidly pre-specified cluster count. %Unlike our proposed GSFC model, FSSC cannot autonomously prune redundant prototypes, and its reliance on superpixels limits its generalizability to raw, non-image datasets.  

Another interesting idea is considered in \cite{Sparse_Fuzzy_In_Features,Sparse_Features_KBS,Sparse_Features_Neurocomputing} where sparsity is investigated in terms of features. For each high-dimensional  data point $\mathbf{x}_i$ only a small number of its features  should dictate the clustering placement, and hence the work focuses on optimizing the feature weighing mechanism.    Consequently, these works do not resolve the fundamental ambiguity of dense fuzzy assignments, nor does it organically shape membership boundaries or prune redundant clusters (again, the true number of clusters is required).

To summarize, the key limitations of the aforementioned works are: \begin{itemize}
    \item \textbf{Limited sparsity inclusion:} Sparsity is typically utilized to filter out noisy points or to select relevant features. 
    \item \textbf{Strong A Priori Assumptions:} All cited works require the exact number of clusters to be known a  priori.
    \item \textbf{Optimization Limitations:} Some works are also restricted to specific fuzzifier values. 
\end{itemize}

\noindent \textbf{Remark:}
It is important to emphasize that the proposed framework does not necessarily compete with or seek to invalidate these prior schemes; rather, they target fundamentally different issues of the clustering framework. Because the previous approaches operate orthogonally to the currently proposed framework, the latter could  work in conjunction with, sparse feature selection or subspace learning algorithms to utilize  multi-level sparse representations produced by the aforementioned methods.

In addition to fuzzy clustering, sparsity can improve the performance of both possibilistic~\cite{K18,SparsePosibilisticGeneral1,SparsePosibilisticGeneral2,SparsePosibilisticGeneral3} and probabilistic~\cite{SparseProbabilisticGeneral1,SparseProbabilisticGeneral2} clustering. For example, \cite{K18}
 develops highly effective algorithms for line segment identification based on probabilistic and possibilistic sparse clustering.  
 
The main issue of \cite{K18} is that the possibilistic framework does not enforce the membership sum constraint, i.e.: 
$$
\sum \text{membership weight} (\mathbf{x}) =1 \quad \forall \text{ data point } \mathbf{x}
$$
Because memberships are entirely independent in the possibilistic rationale, the clusters do not mathematically compete for data points. Without this competition, multiple cluster representatives can easily collapse onto the same dense region of data, requiring highly sensitive hyperparameter tuning to keep   nearby clusters separated.
This issue is addressed by \cite{S_and_K21}.  However, the proposed iterative algorithm  relies on \textbf{specific  distribution models} to define the \textbf{spread} of each cluster. This parametric assumption can be heavily limiting in applications where the data is not Gaussian-like or where distinct clusters exhibit vastly different spatial variances.

Our work aims to address the weaknesses of both \cite{K18} and \cite{S_and_K21} by combining the concept of  sparsity with fuzzy clustering. By doing so, the resulting framework strictly enforces the competitive membership sum-to-one constraint, while it remains model-agnostic, allowing it to successfully address general, non-Gaussian data structures.
The main contributions of this work are summarized as follows:
\begin{itemize}
    \item \textbf{A Novel General Sparse Fuzzy Clustering (GSFC) Framework:} We introduce a continuous sparsity penalty term into the traditional fuzzy objective. This mathematically enforced sparsity enables GSFC to autonomously discover the true number of clusters by organically pruning redundant clusters from an over-initialized state. The solution of the resulting CFO problem is carried out through the utilization of a novel bisection-driven optimization mechanism.
  To overcome the extreme sensitivity of dense fuzzy assignments to outliers, we integrate a "dummy" noise cluster into the GSFC optimization. This mechanism safely absorbs noisy data points, preventing them from  affecting the positions of the cluster representatives.

   \item \textbf{Improvement over Dense Fuzzy  Methods:} Through comprehensive experiments on both synthetic toy datasets, real-world tabular data, and HyperSpectral Images (HSIs), we demonstrate that the enforced sparsity allows GSFC to match or exceed the performance of traditional Fuzzy C-Means (FCM), while  providing the true cluster cardinality. Furthermore, our experimental results reveal exceptional stability across random initializations, as over-initializing the candidate cluster representatives pool, followed by a dynamically  pruning procedure, ensures thorough coverage of the data space and prevents convergence traps caused by poorly placed initial cluster centers.

    \item \textbf{State-of-the-Art Line Segment Extraction:} Apart from the use of the proposed framework for the case of compact clusters, we demonstrate the use of  GSFC  to the complex geometrical task of line segment identification. The model's ability to drive trace memberships to exact zero allows it to significantly outperform alternative soft clustering algorithms in identifying intersecting physical segments.
  \end{itemize}

The rest of this paper is organized as follows. The GSFC framework for point representatives is presented in section II, and an extension to the special case of elongated clusters is provided in section III. A wide range of experiments in  both synthetic and real-world data is provided in section IV, and section V concludes.

 \section{General Sparse Fuzzy Clustering (GSFC)}
\subsection{Objective Function and Lagrangian}
Let $\mathcal{X} = \{\mathbf{x}_1, \dots, \mathbf{x}_N\} \subset \mathbb{R}^d$ be a set of $N$ data points, and $\Theta = \{\boldsymbol{\theta}_1, \dots, \boldsymbol{\theta}_m\}$ be the set of $m$ cluster representatives, corresponding to the clusters $C_1,\cdots,C_m$, respectively.  
We consider first the case where the data points form compact clusters and, as a consequence, the cluster representatives $\boldsymbol{\theta}_s$, $s=1,\cdots,m$, are points in $R^d$ (in Section III, it is shown how GSFC can be extended to handle the case of elongated clusters corresponding to line-segments).
The objective/cost function of General Sparse Fuzzy Clustering (GSFC) is defined as:
\begin{align}
    J_{\text{GSFC}}(\mathbf{U}, \Theta) &= \sum_{i=1}^{N} \sum_{j=1}^{m} u_{ij}^q d(\mathbf{x}_i, \boldsymbol{\theta}_j) + s \sum_{i=1}^{N} \sum_{j=1}^{m} u_{ij}^p = \sum_{i=1}^{N} \sum_{j=1}^{m} u_{ij}^q d(\mathbf{x}_i, \boldsymbol{\theta}_j) + s\sum_{i=1}^N ||\mathbf{u}_i||_p^p,
\end{align}
 where $d(\mathbf{x}_i,\boldsymbol{\theta}_j)$ is the euclidean distance of the i-th data point from the j-th representative, $\mathbf{U} = [u_{ij} ] \in \mathbb{R}^{N \times m} $ is the membership
matrix, where $u_{ij}$ is the degree of belongingness of the i-th data point to the j-th cluster, and $q(>1)$ is the so-called fuzzifier parameter. Also,
$0 < p < 1$ is the sparsity exponent (hence, $||\mathbf{v}||_p$ is the quasinorm of vector $\mathbf{v}$), $s > 0$ controls sparsity strength.

The first term in (1) is the traditional fuzzy objective and the second  is  the \textbf{sparsity}-imposing term. In the general fuzzy clustering framework, optimization of the cost function is subject to the following constraints:
\begin{align}
    \sum_{j=1}^{m} u_{ij} = 1, \quad &\forall i \in \{1, \dots, N\} \label{eq:sum_constraint} \\
    u_{ij} \in [0, 1], \quad &\forall i \in \{1, \dots, N\}, \, j \in \{1, \dots, m\},
\end{align}
To minimize $J_{\text{GSFC}}$ with respect to $\mathbf{U}$ under constraint~\eqref{eq:sum_constraint}, we construct the Lagrangian function:

\begin{align}\label{eq:obj_lagrangian_def}
    \mathcal{L}(\mathbf{U}, \Theta, \boldsymbol{\lambda}) &= \sum_{i=1}^{N} \sum_{j=1}^{m} u_{ij}^q d(\mathbf{x}_i, \boldsymbol{\theta}_j) + s \sum_{i=1}^{N} \sum_{j=1}^{m} u_{ij}^p \nonumber\\
    & \hspace{2cm}- \sum_{i=1}^{N} \lambda_i \left( \sum_{j=1}^{m} u_{ij} - 1 \right)
\end{align}

\noindent where $\boldsymbol{\lambda} = [\lambda_1, \dots, \lambda_N]^T$ is the vector of Lagrange multipliers. Setting the partial derivative with respect to $u_{rs}$ (for a specific data point $\mathbf{x}_r$ and a specific cluster $\boldsymbol{\theta}_s$) to zero yields the first-order stationarity condition:

\begin{equation}
    \frac{\partial \mathcal{L}}{\partial u_{rs}} = q d_{rs} u_{rs}^{q-1} + s \cdot p u_{rs}^{p-1} - \lambda_r = 0 \label{eq:stationarity}
\end{equation}

\noindent where $d_{rs} \equiv d(\mathbf{x}_r, \boldsymbol{\theta}_s)$.

\noindent \textbf{Remark:}
Unlike the classic (i.e., without the  sparsity loss term)  fuzzy clustering frameworks e.g., FCM, in this framework we can not extract closed-form solution for $u_{rs}$ from eq. \eqref{eq:stationarity}, due to the simultaneous presence of $u_{rs}^{q-1}$ and $u_{rs}^{p-1}$.
% =========================================================
\subsection{Determining Sparse Memberhips $u_{rs}$ using the Bisection Method}
 Here we will discuss an optimization procedure for inferring the membership assignments for a data point $\mathbf{x}_r$ from eq. \eqref{eq:stationarity}. 
Focusing on a specific cluster $s$ we need to solve $$q d_{rs} u_{rs}^{q-1} + s p u_{rs}^{p-1} - \lambda_r = 0.$$ 
To eliminate the negative exponent ($p-1 < 0$), we multiply both sides of the equation by $u_{rs}^{1-p}$, yielding:
\begin{equation}
q d_{rs} u_{rs}^{q-p} + s p - \lambda_r u_{rs}^{1-p} = 0
\end{equation}
We introduce a substitution variable $v = u_{rs}^{1-p}$ (implying $u_{rs} = v^{\frac{1}{1-p}}$) and define the exponent scalar $a = \frac{q-1}{1-p} > 0$. This implies that
\begin{equation}q-p = (a+1)(1-p),\end{equation} 
allowing us to rewrite $u_{rs}^{q-p}$ as $v^{a+1}$. Eq. (\ref{eq:stationarity}) elegantly reduces to finding the roots of the generalized function:
\begin{equation}\label{eq:stationarity_reduction}
f(v) \equiv q d_{rs} v^{a+1} - \lambda_r v + s p = 0\end{equation}
To determine if valid non-negative memberships exist, we analyze the graph of $f(v)$. At $v=0$, $f(0) = sp > 0$. Taking the first derivative with respect to $v$ and setting it to zero yields the location of the local extremum:
\begin{align}\label{eq:v_min_def}
f'(v) &= (a+1) q d_{rs} v^a - \lambda_r = 0 \implies \nonumber \\
v_{min} & = \left( \frac{\lambda_r}{(a+1) q d_{rs}} \right)^{\frac{1}{a}}\end{align}

Because the second derivative $f''(v) = (a+1) a q d_{rs} v^{a-1}$ is strictly positive for all $v > 0$, $v_{min}$ represents the unique global minimum of $f$ for $v > 0$. The existence of valid membership roots depends entirely on the value of $f(v_{min})$. We consider two cases: 

 \noindent \textbf{Sparse Pruning} ($f(v_{min}) > 0$): The minimum sits above the x-axis, meaning no positive real roots exist, which in turn means that the  data point is too far from the cluster, relative to the dual variable $\lambda_r$. In this case, the optimal assignment is driven to exact zero: $u_{rs} = 0$.

\noindent \textbf{Active Condition} ($f(v_{min}) \le 0$): The function dips below the x-axis, producing, in general, two positive real roots. 

When $v < v_{\min}$, the derivative $f'(v)$ is strictly negative, which translates to a negative second derivative of the GSFC objective ($\frac{\partial^2 \mathcal{L}}{\partial u_{rs}^2} < 0$). In this region, the local objective segment is concave. Therefore, the stationary point $v_1$ ( the smaller solution of $\frac{\partial \mathcal{L}}{\partial u_{rs}} = 0$) is a local maximum. 

Conversely, when $v > v_{\min}$, $f'(v)$ becomes strictly positive. This flips the second derivative of the initial objective to be strictly positive as well ($\frac{\partial^2 \mathcal{L}}{\partial u_{rs}^2} > 0$), making the local objective strictly convex in that region. %On a convex curve, any point where the derivative is $0$ satisfies the second-order sufficiency condition, uniquely designating it as a local minimum. 
Therefore, $v_2$, the larger solution of $\frac{ \partial {\cal L} }{ \partial u_{rs} }=0$, is a local minimum of $\cal L$ with respect to $u_{rs}$.
Hence, the optimal $v^*$ lies in the interval $(v_{\min},1)$
\subsubsection{General Newton-Raphson+Bisection Solver}
Because $f(v) = 0$ lacks a closed-form expression for arbitrary  values of the exponent $a$, the active condition ($f(v_{\min}) \le 0$) requires a two-level numerical solver to determine $u_{r1},...,u_{rm}$ for a specific data point $\mathbf{x}_r$: an outer bisection search for the Lagrange multiplier $\lambda_r$, and an inner Newton-Raphson (NR) solver to compute $v^*$  at every (outer) bisection step,  corresponding to a specific $\lambda_r$ value.  The NR solver is utilized $m$ times, one for each $u_{rs}, s=1,\cdots,m$. For a specific $s$, it is initialized to the right of the corresponding $v_{\min}$.

For a given candidate dual variable $\lambda_{\text{mid}}$ generated by the outer bisection search, the NR inner loop refines the root iteratively via:
\begin{align}
v^{(t+1)} &= v^{(t)} - \frac{f(v^{(t)})}{f'(v^{(t)})} \nonumber \\
&= v^{(t)} - \frac{q d_{rs} (v^{(t)})^{a+1} - \lambda_{\text{mid}} v^{(t)} + s p}{(a+1) q d_{rs} (v^{(t)})^a - \lambda_{\text{mid}}}
\end{align}
Upon convergence, the active sparse membership is recovered via $$u_{rs} = (v^*)^{\frac{1}{1-p}}.$$ 
Having applied the above procedure $m$ times for the specific point $\mathbf{x}_r$, the
 outer bisection loop evaluates the partition residual $\Phi(\lambda_{\text{mid}}) = \sum_{s=1}^m u_{rs} - 1$, updating the search bounds until $|\Phi| < \epsilon_{bis}$, , where $\epsilon_{bis}$ is a small user-defined parameter.

One important detail is how  the search interval for the bisection method $(\lambda_{\min},\lambda_{\max})$ is configured.
A trivial lower bound can be configured by setting $\lambda_{\min}=0$. Then $f(v)$ is a sum of positive numbers and hence, $f(v_{\min})>0$.   This triggers the Sparse Pruning condition for every single cluster, forcing $u_{rs} = 0$ for all $s$. Hence the partition residual is $\Phi(0)=0-1=-1$.

To get an upper bound, we
start from the general equation 
$$q d_{rs} u_{rs}^{q-1} + s p u_{rs}^{p-1} = \lambda_r.$$
If we substitute $u_{rs} = 1$ and use the minimal distance $ \min_s(d_{rs})$, the exponents disappear ($1^{q-1} = 1$ and $1^{p-1} = 1$), leaving us with the exact theoretical upper bound:\begin{equation}\label{eq:lambda_max_def}\lambda_{\text{max}} = q \min_s(d_{rs}) + s p\end{equation}
Since $u_{rs}$ will be equal to $1$ for one cluster and positive for the rest, it will hold $\Phi(\lambda_{\max}) \geq 0$. The above procedure is described in its detail in Algorithm  1.
\subsubsection{Special  Pathways Leading to Analytic Solutions for \eqref{eq:stationarity}}
The computational overhead of running an inner NR solver inside the bisection loop   can be bypassed for specific choices of values of the   parameter pairs $(p, q)$ that collapse $f(v) = 0$ into exact closed-form algebraic solutions. Specifically, setting $p = 0.5$ (as in \cite{K18}) yields two primary pathways:
\begin{itemize}
    \item \textbf{Pathway 1 ($q = 1.5, p = 0.5 \implies a = 1$):} Reduces eq. (8) to a quadratic equation $1.5 d_{rs} v^2 - \lambda_r v + 0.5 s = 0$, yielding the direct root formula:
    \begin{equation}\label{eq:pathway_updates}
    u_{rs}(\lambda_r) = \left( \frac{\lambda_r + \sqrt{\lambda_r^2 - 3 s d_{rs}}}{3 d_{rs}} \right)^2 \quad \text{for } \lambda_r \ge \sqrt{3 s d_{rs}}
    \end{equation}
    Otherwise, $u_{rs}$ is set to $0$.
    \item \textbf{Pathway 2 ($q = 2.0, p = 0.5 \implies a = 2$):} Reduces to a depressed cubic equation $2.0 d_{rs} v^3 - \lambda_r v + 0.5 s = 0$, solved analytically via Viète's trigonometric roots:
    \begin{equation}
    u_{rs}(\lambda_r) = 2 \sqrt{\frac{\lambda_r}{6 d_{rs}}} \cos \left( \frac{1}{3} \arccos \left( -\frac{s}{4} \sqrt{\frac{54 d_{rs}^3}{\lambda_r^3}} \right) \right)
    \end{equation}
\end{itemize}
In the sequel we will experimentally showcase that both of these pathways can produce very strong performance, across diverse datasets.
\begin{algorithm}[H]
\caption{Subroutine: Sparse Membership Vector Evaluation for Sample $\mathbf{x}_r$ (Invoked iteratively per data point within GSFC)}
\label{alg:membership_update}
\begin{algorithmic}[1]
%    \STATE \textbf{Procedure} $\textsc{ComputeSparseMembership}(\mathbf{d}_r, s, p, q, \epsilon_{bis})$
    \STATE \textbf{Input:} Distance vector $\mathbf{d}_r = [d_{r1}, \dots, d_{rm}]^T$ for a single sample $\mathbf{x}_r$, sparsity factor $s > 0$, parameters $(p, q)$, bisection tolerance $\epsilon_{bis} > 0$.
    \STATE \textbf{Output:} Sparse membership vector $\mathbf{u}_r = [u_{r1}, \dots, u_{rm}]^T$ and optimal Lagrange multiplier $\lambda_r^*$.

    \STATE Set bounds $\lambda_{\text{min}} \leftarrow 0$ and  $\lambda_{\text{max}} = q \min_s(d_{rs}) + s p$ 
    \STATE Set $\lambda_{\text{low}} \leftarrow \lambda_{\text{min}}$, $\lambda_{\text{high}} \leftarrow \lambda_{\text{max}}$
    \WHILE{$(\lambda_{\text{high}} - \lambda_{\text{low}}) > \epsilon_{bis}$}
        \STATE $\lambda_{\text{mid}} \leftarrow \frac{\lambda_{\text{low}} + \lambda_{\text{high}}}{2}$
        \FOR{$s = 1 \dots m$}
            \STATE Compute $a = \frac{q-1}{1-p}$ and $v_{\min} = \left( \frac{\lambda_{\text{mid}}}{(a+1) q d_{rs}} \right)^{\frac{1}{a}}$
            \STATE Evaluate active condition: $f(v_{\min}) = q d_{rs} (v_{\min})^{a+1} - \lambda_{\text{mid}} v_{\min} + s p$
            \IF{$f(v_{\min}) > 0$}
                \STATE $u_{rs} \leftarrow 0$ \COMMENT{\textit{Sparse Pruning}}
            \ELSE
                \STATE \textbf{Find root $v^* \ge v_{\min}$ of $f(v) = 0$} \COMMENT{\textit{Via closed-form pathways or NR}}
                \STATE $u_{rs} \leftarrow (v^*)^{\frac{1}{1-p}}$
            \ENDIF
        \ENDFOR
        \STATE Evaluate partition residual: $\Phi \leftarrow \sum_{s=1}^{m} u_{rs} - 1$
        \IF{$|\Phi| < \epsilon_{bis}$}
            \STATE \textbf{break} \COMMENT{\textit{Convergence achieved}}
        \ELSIF{$\Phi > 0$}
            \STATE $\lambda_{\text{high}} \leftarrow \lambda_{\text{mid}}$ \COMMENT{\textit{Sum too large $\implies$ decrease $\lambda_r$}}
        \ELSE
            \STATE $\lambda_{\text{low}} \leftarrow \lambda_{\text{mid}}$ \COMMENT{\textit{Sum too small $\implies$ increase $\lambda_r$}}
        \ENDIF
    \ENDWHILE
    \STATE Set $\lambda_r^* \leftarrow \lambda_{\text{mid}}$ and \textbf{return} $\mathbf{u}_r = [u_{r1}, \dots, u_{rm}]^T$
    \STATE \textbf{End Subroutine}
\end{algorithmic}
\end{algorithm}

\subsubsection{Validity of the Membership Bounds}
A critical mathematical requirement for fuzzy clustering is that the resulting $u_{rs}$ memberships strictly lie within the valid   interval $  [0, 1]$ for the  $\lambda_r$ value outputted by algorithm 1. In the proposed framework, these bounds are strictly enforced through two distinct mechanisms:

\textbf{Non-negativity ($u_{rs} \ge 0$):} The lower bound is guaranteed directly by the closed-form derivations in both evaluation pathways. In Pathway 1, the thresholding condition explicitly sets $u_{rs} = 0$ if $\lambda_r < \sqrt{3sd_{rs}}$. If this condition is met, the active formula evaluates to a squared real number. Similarly, in Pathway 2, the cubic discriminant check sets $u_{rs} = 0$ if $\Delta_{rs} > 0$, while Vi\`ete's formula yields a real root $v^*$ that is subsequently squared. Thus, for any evaluated $\lambda_r$, it is mathematically guaranteed that $u_{rs}(\lambda_r) \ge 0$.

For the general case with arbitrary exponents  ($q > 1, 0 < p < 1$)  non-negativity is preserved. Under the sparse pruning condition ($f(v_{\min}) > 0$), the membership is directly assigned to $u_{rs} = 0$. When the active condition holds ($f(v_{\min}) \le 0$), the critical point $$v_{\min} = \left(\frac{\lambda_r}{(a+1)qd_{rs}}\right)^{1/a}$$ is strictly positive since $\lambda_r, d_{rs}, q, a > 0$. As we have shown, in that case, the objective's local minimum  $v^*$ resides on the  branch $ (v_{\min}, 1)$. As a result, the inverse power transformation $u_{rs} = (v^*)^{\frac{1}{1-p}}$ is a positive real power of a strictly positive real number (as $1-p > 0$). Thus, $u_{rs} > 0$ strictly in the active regime, guaranteeing that $u_{rs} \ge 0$ holds universally for any valid $(p, q)$ configuration.\color{black}

\textbf{Upper Bound ($u_{rs} \le 1$):} The upper bound is not explicitly restricted by the individual membership formulas, but is rather implicitly guaranteed by the bisection search regardless of the optimization pathway. Because the algorithm terminates only when the fuzzy partition constraint $\sum_{s=1}^{m} u_{rs}(\lambda_r^*) = 1$ is  satisfied, and because all individual memberships are  guaranteed to be non-negative, the fundamental properties of addition dictate that no single membership can exceed $1$. Therefore, at the optimal dual variable $\lambda_r^*$, the constraint $u_{rs} \in [0, 1]$ is strictly satisfied.
% =========================================================
\subsection{Determination of the Optimal Number of Clusters.}
To automatically determine the optimal number of clusters $m$ and safeguard against numerical instabilities, the iterative optimization loop incorporates two dynamic post-processing procedures: cardinality-based elimination and distance-based merging.

\subsubsection{Cluster Elimination (Cardinality-Based Pruning)}
Because of the sparsity imposition, some clusters may lose membership support,  that is, almost all data points have very low membership values for these clusters, which, in turn, it translates to the the fact that these clusters are "almost empty".  To identify whether a specific cluster $C_j$ is  underpopulated or isolated, we compute the effective cardinality $N_j$ (total accumulated membership):
\begin{equation}
    N_j = \sum_{r=1}^{N} u_{rj}
\end{equation}
Then $C_j$ is declared as degenerate and is  eliminated if its effective cardinality falls below a user-defined threshold $\epsilon_{\text{elim}} \ge 0$. % (typically $\epsilon_{\text{elim}} \in [0, 1]$):
%\begin{equation}
%    \text{If } N_j < \epsilon_{\text{elim}}, \quad \text{remove } \mathbf{v}_j \text{ from } V \text{ and drop column } j \text{ from } U
%\end{equation}

This pruning mechanism serves two key roles: it allows the algorithm to be safely over-initialized with a large candidate cluster count $m^{(0)}$, and it also prevents division-by-zero singularities during prototype update steps (which scale inversely with $N_j$).  

\subsubsection{Cluster Merging (Distance-Based Coalescence)}
Over-initialization or overlapping clusters can cause multiple cluster representatives to converge onto the same area of a single physical cluster (erroneously splitting it to two or more clusters). To eliminate redundant representatives (and, at the same time, the clusters they are associated with), a pairwise distance check is conducted across all active clusters. For any pair of representatives $(\boldsymbol{\theta}_j, \boldsymbol{\theta}_k)$, if their Euclidean distance falls below a merge threshold $\epsilon_{\text{gap}} > 0$:
\begin{equation}
    d(\boldsymbol{\theta}_j,  \boldsymbol{\theta}_k) = \|\boldsymbol{\theta}_j - \boldsymbol{\theta}_k\|_2 < \epsilon_{\text{gap}}
\end{equation}
the two clusters are collapsed into a single prototype $\boldsymbol{\theta}_{\text{new}}$ using a cardinality-weighted average:
\begin{equation}
    \boldsymbol{\theta}_{\text{new}} = \frac{N_j \boldsymbol{\theta}_j + N_k \boldsymbol{\theta}_k}{N_j + N_k}
\end{equation}
Once the prototypes are merged, columns $j$ and $k$ in the membership matrix $U$ are removed. The exact optimal memberships and distances for the unified prototype $\theta_{\text{new}}$ are naturally recomputed during the evaluation phase of the subsequent iteration.
Note that  directional or manifold prototype geometries (such as line segments) can incorporate angular alignment metrics into the merge criterion (an example, based on \cite{K18} is provided in the next section). 
\subsection{Cluster Representative Update}
Once the optimal sparse membership matrix $\mathbf{U}^{(t)}$ is computed (where $t$ denotes the current optimization loop iteration) , the objective function $J_{\text{GSFC}}$ of eq. (1) decouples across the active clusters, allowing each representative to be updated independently by solving $\boldsymbol{\theta}_j^{(t+1)} = \arg\min_{\boldsymbol{\theta}_j} \sum_{r=1}^{N} u_{rj}^q(t) d(\mathbf{x}_r, \boldsymbol{\theta}_j)$. In the present case of point representatives, the  updating equation  takes the following closed-form expression:
$$\boldsymbol{\theta}_j^{(t+1)} = \frac{ \sum_{r=1}^N u_{rj}^q(t) \mathbf{x}_r }{ \sum_{r=1}^N u_{rj}^q(t) }$$
The GSFC framework is described analytically in Algorithm 2.

\begin{algorithm}[H]
\caption{General Sparse Fuzzy Clustering (GSFC)}
\label{alg:gsfc_main}
\begin{algorithmic}[1]
    \STATE \textbf{Input:} Data $\mathcal{X} = \{\mathbf{x}_1, \dots, \mathbf{x}_N\}$, initial prototype count $m^{(0)}$, hyperparameters $(q, p, s)$, thresholds $(\epsilon_{\text{elim}}, \epsilon_{\text{gap}}, \epsilon_{\text{conv}})$, small constant $\epsilon_d = 10^{-12}$, max iterations $t_{\max}$.
    
    \STATE \textbf{Output:} Optimized cluster prototypes $\Theta = \{\boldsymbol{\theta}_1, \dots, \boldsymbol{\theta}_m\}$ and sparse membership matrix $U \in \mathbb{R}^{N \times m}$.
    
    \STATE \textbf{Initialize:} Overestimated prototypes $\Theta^{(0)} = \{\boldsymbol{\theta}_1, \dots, \boldsymbol{\theta}_{m^{(0)}}\}$, iteration counter $t \leftarrow 0$.
    
    \REPEAT
        \STATE \textbf{Distance Calculation:} For all $r \in \{1,\dots,N\}$ and $j \in \{1,\dots,m\}$, compute regularized distances $d_{rj} = \max\left(d(\mathbf{x}_r, \boldsymbol{\theta}_j), \epsilon_d\right)$.
        
        \STATE \textbf{Membership Update:} For each point $\mathbf{x}_r$, determine $\lambda_r^*$  and compute sparse vector $\mathbf{u}_r(t) = [u_{r1}, \dots, u_{rm}]^T$  calling Algorithm 1$\left(\mathbf{x}_r,[d_{r1},\cdots,d_{rm}],s,q,p\right)$. 
        \STATE \textbf{Cluster Elimination:} Compute effective cardinality $N_j = \sum_{r=1}^N u_{rj}(t)$ for all $j$. For any cluster with $N_j < \epsilon_{\text{elim}}$, remove prototype $\boldsymbol{\theta}_j$ and update total count $m$.
        
        \STATE \textbf{Cluster Update:} For each active cluster $j \in \{1, \dots, m\}$, update parameters:
        \[
            \boldsymbol{\theta}_j^{(t+1)} = \arg\min_{\boldsymbol{\theta}_j} \sum_{r=1}^{N} u_{rj}^q(t) \, d(\mathbf{x}_r, \boldsymbol{\theta}_j)
        \]
        
        \STATE \textbf{Cluster Merging:} Identify pairs satisfying $d(\boldsymbol{\theta}_j, \boldsymbol{\theta}_k) \le \epsilon_{\text{gap}}$. Merge into $\boldsymbol{\theta}_{\text{new}} = \frac{N_j \boldsymbol{\theta}_j + N_k \boldsymbol{\theta}_k}{N_j + N_k}$.
        
        \STATE $t \leftarrow t + 1$
    \UNTIL{$\|\Theta^{(t)} - \Theta^{(t-1)}\|_F \le \epsilon_{\text{conv}}$ \textbf{or} $t \ge t_{\max}$}
\end{algorithmic}
\end{algorithm}

\subsection{Handling Outliers via a Virtual Noise Sink}
To prevent distant outliers from acting as heavy anchors that drag prototypes out of place under the strict partition constraint ($\sum_{j=1}^{m} u_{rj} = 1$), we integrate a virtual noise sink governed by a constant distance threshold, $d_{\text{noise}}$. Specifically, a dummy cluster is appended at a fixed distance $d_{\text{noise}}$ from all prototypes. During membership evaluation, any data point exceeding this threshold from the active prototypes assigns its membership entirely to the dummy cluster. This forces its membership to the true structural segments to zero, completely excluding background outliers from prototype updates while explicitly flagging them as noise during final evaluation.  
\subsection{Visualizing Pruning Procedure}
\begin{figure}
    \centering
    \includegraphics[width=\linewidth]{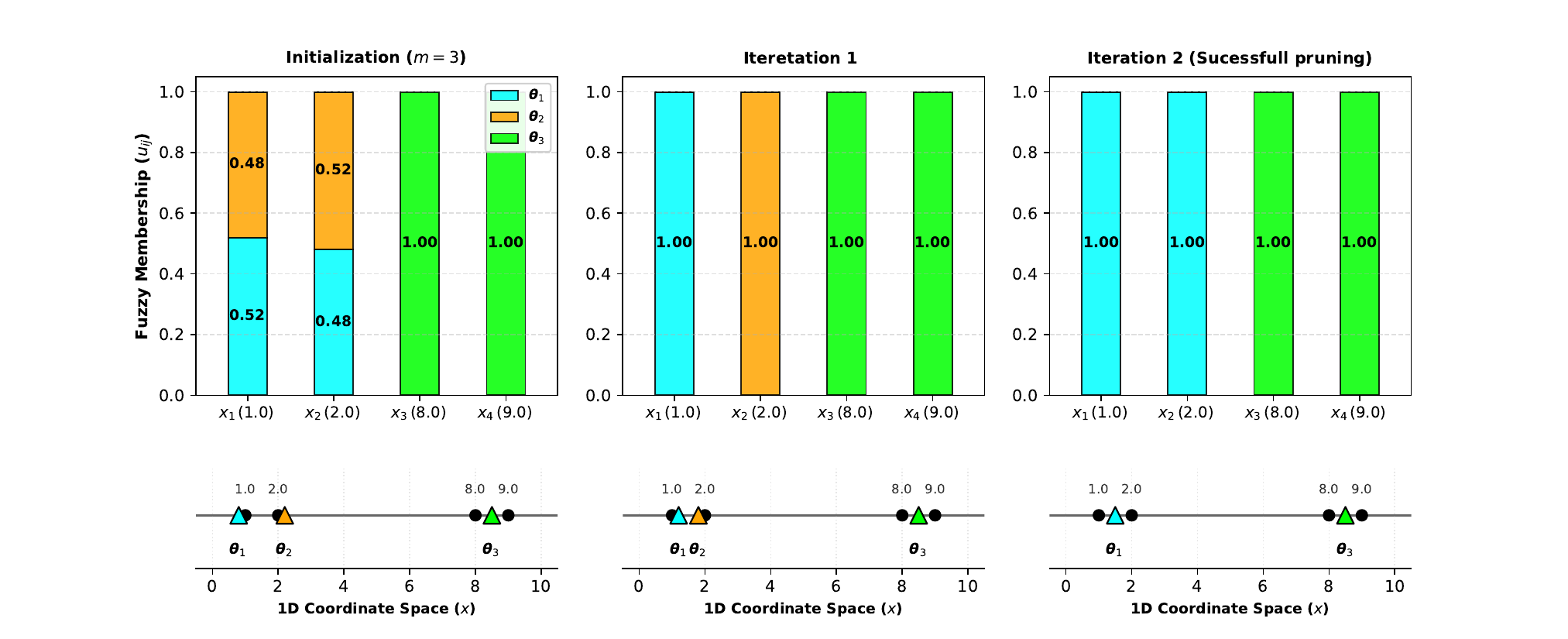}
    \caption{Visualization of GSFC's pruning process}
    \label{fig:pruning_viz}
\end{figure}

To better understand how the algorithm eliminates redundant prototypes we constructed a toy example with $4$ 1D points $x_1,x_2,x_3,x_4$ positioned at $1,2,8,9$ and initialized the GSFC algorithm (for q=1.5 and p=0.5) with $3$ cluster representatives $\boldsymbol{\theta}_1=0.8,\boldsymbol{\theta}_2=2.2$, and $\boldsymbol{\theta}_3=8.5$. The merging threshold is set to $\epsilon_{\text{gap}}=1$.  The operation of GSFC for 3 optimization iterations is depicted in Fig. \ref{fig:pruning_viz}.  The first optimization pathway was considered without a noise cluster.

Initially, both $x_1$ and $x_2$ are  very close to prototypes $1$ and $2$ leading to genuine fuzzy sharing with memberships of approximately $0.52$ and $0.48$. Due to the sparsity penalty $u_{1,3}=u_{2,3}=0$. Likewise $x_3,x_4$ are only assigned to $\boldsymbol{\theta}_3$ due to  sparsity. 

In Step 1, as $\boldsymbol{\theta}_1$ and $\boldsymbol{\theta}_2$ shift inward to $1.2$ and $1.8$, they localize tightly around their nearest individual points ($d_{1,1}=0.04$ versus $d_{1,2}=0.64$ and vice-versa). Because GSFC enforces a distance-dependent activation threshold (see the condition of eq. (\ref{eq:pathway_updates})) representative $\boldsymbol{\theta}_2$ is "excluded" from the clusters connected to $x_1$ and vice-versa (i.e. the membership becomes $0$). Consequently, the memberships snap to compact assignments of $1.00$ for each point's local prototype, demonstrating how the algorithm eliminates background fuzziness when prototypes become locally dominant.

Finally, in Step 2, despite the sharp local partitioning in Step 1, the spatial proximity condition  identifies $\boldsymbol{\theta}_1$ and $\boldsymbol{\theta}_2$ as redundant representations of the same dense data region. They fuse into a single cluster representative at $1.50$, restoring a unified representation of $\{x_1, x_2\}$ and pruning the cluster count from $3$ to the true $m^*=2$.

\section{GSFC for Elongated Clusters}\label{sec:line_segment_detection}Adapting GSFC for the case of elongated clusters requires redefining representatives as finite segments $\boldsymbol{\theta}_j$ bounded by endpoints $\mathbf{a}_j$ and $\mathbf{b}_j$ along infinite carrier lines $H_j$ ($\mathbf{w}_j^T \mathbf{x} + w_{j0} = 0$).  
A notable example where the elongated cluster case is useful is in line segment detection in edge-pixel images that contain linear elements. In this case, the data points are the image edge pixels, and the aim is to group the pixels that form a certain line to the same cluster.

Here, we briefly review how clustering mechanisms can be utilized for this application, based on the rigorous study of \cite{K18}.  The core sparsity and bisection mechanics remain unchanged, but pruning and merging rules are specialized for linear structures.\subsection{Segment-to-Point Distance Metric}
The distance $d(\mathbf{x}_i, \boldsymbol{\theta}_j)$ combines the squared perpendicular distance to the infinite carrier line $H_j: \mathbf{w}_j^T \mathbf{x}_i+{w}_{j0}=0 $ (first term in eq. (17) below) with the distance of the projection of $\mathbf{x}_i$ 
on $H_j$,
$\mathbf{x}_i^j$, from the finite segment,  which is completely defined by its two end points $\mathbf{a}_j$ and $\mathbf{b}_j$, i.e., $\boldsymbol{\theta}_j=[\mathbf{a}_j, \mathbf{b}_j]$ (second term in (17) below).
\begin{equation}
d(\mathbf{x}_i, \boldsymbol{\theta}_j) = \frac{(\mathbf{w}_j^T \mathbf{x}_i + w_{j0})^2}{\Vert{}\mathbf{w}_j\Vert{}^2} + d(\mathbf{x}_i^j, \boldsymbol{\theta}_j)
\end{equation}
where $\mathbf{x}_i^j = \mathbf{x}_i - \frac{\mathbf{w}_j^T \mathbf{x}_i + w_{j0}}{\Vert{}\mathbf{w}_j\Vert{}^2}\mathbf{w}_j$.
If $\mathbf{x}_i^j$ falls outside $\boldsymbol{\theta}_j$ (that is, outside the interval on $H_j$ with endpoints $\mathbf{a}_j$ and $\mathbf{b}_j$), $d(\mathbf{x}_i^j, \boldsymbol{\theta}_j)=\min(d(\mathbf{x}_i^j,\mathbf{a}_j),d(\mathbf{x}_i^j,\mathbf{b}_j)) $. %evaluates to the minimum squared distance to endpoints $\mathbf{a}_j$ or $\mathbf{b}_j$.
Otherwise, $d(\mathbf{x}_i^j,\boldsymbol{\theta}_j)=0$.
\subsection{Representative Update Rule}Representatives are updated by solving a weighted eigenvalue problem. Given the weighted mean $$\boldsymbol{\xi}_j = \frac{\sum u_{ij} \mathbf{x}_i}{\sum u_{ij}},$$ 
and covariance matrix
$$\mathbf{A}_j = \sum u_{ij} (\mathbf{x}_i-\boldsymbol{\xi}_j) (\mathbf{x}_i - \boldsymbol{\xi}_j)^T,$$ 
the orientation vector $\mathbf{w}_j$ is the eigenvector corresponding to the smallest eigenvalue of $\mathbf{A}_j$. The intercept is updated as $w_{j0} = \mathbf{w}_j^T \boldsymbol{\xi}_j$. New endpoints $\mathbf{a}_j$ and $\mathbf{b}_j$ are determined by projecting points with significant membership ($u_{ij} > 0.05$) onto $H_j$.
\subsection{Cluster Elimination and Merging}
Standard continuous pruning is replaced by a discrete cardinality requirement: a segment is pruned if it fails to serve as the primary representative for at least two distinct  points ($|S_j| < 2$), where $S_j = \{ \mathbf{x}_i \in \mathcal{X} \mid j = \arg\max_k u_{ik} \}$. To handle redundant clusters, two  line segments are merged if they are (i) collinear ($\arccos(|\mathbf{w}_j^T \mathbf{w}_k|) \le \epsilon_{\text{angle}}$) and (ii) spatially contiguous, meaning their spans overlap or their nearest endpoints lie within a small distance gap $\epsilon_{\text{gap}}$. %When this condition is met, the two clusters coalesce into a single representative spanning the combined len.
\section{Numerical Experiments}
In this section, we conduct a comprehensive empirical evaluation of GSFC across a diverse benchmark suite, comprising two synthetic toy datasets, two real-world tabular benchmarks, three HSIs, and multiple line-segment detection scenarios. The primary objective of these experiments is twofold. First, we aim to validate that introducing  sparsity into the fuzzy membership assignment process yields substantial performance improvements and noise robustness over conventional, non-sparse FCM. Second, we seek to demonstrate the superiority of the proposed framework over competing   sparse clustering algorithms developed for the elongated cluster case.

\begin{center}
    \begin{minipage}{\linewidth}
        \centering
        \captionsetup{type=figure} % Tells subcaption this block behaves as a figure
        
        \subcaptionbox{Experiment A \label{fig:ari_noise}}{%
            \includegraphics[width=0.48\linewidth]{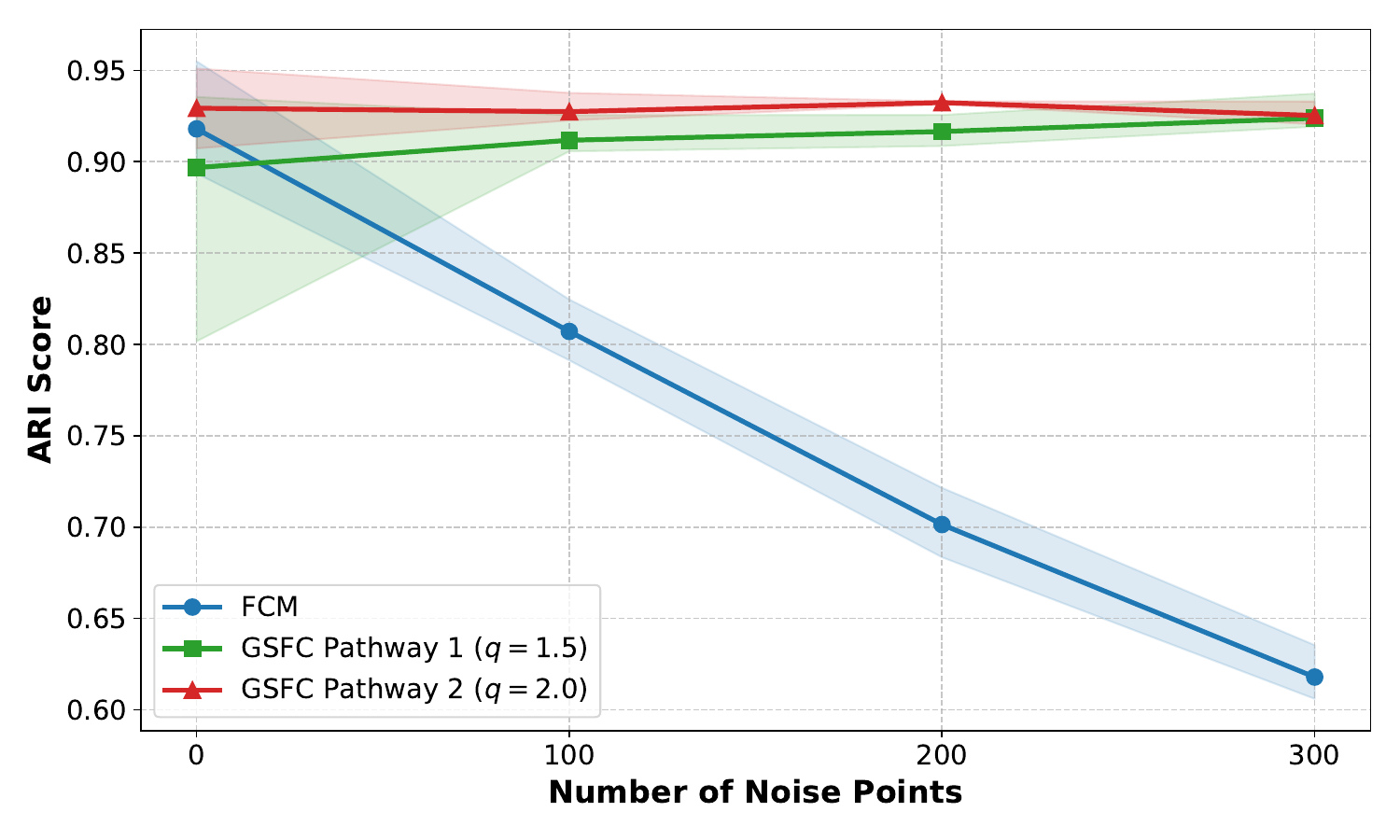}%
        }%
        \hfill
        \subcaptionbox{Experiment B\label{fig:ari_laplace}}{%
            \includegraphics[width=0.48\linewidth]{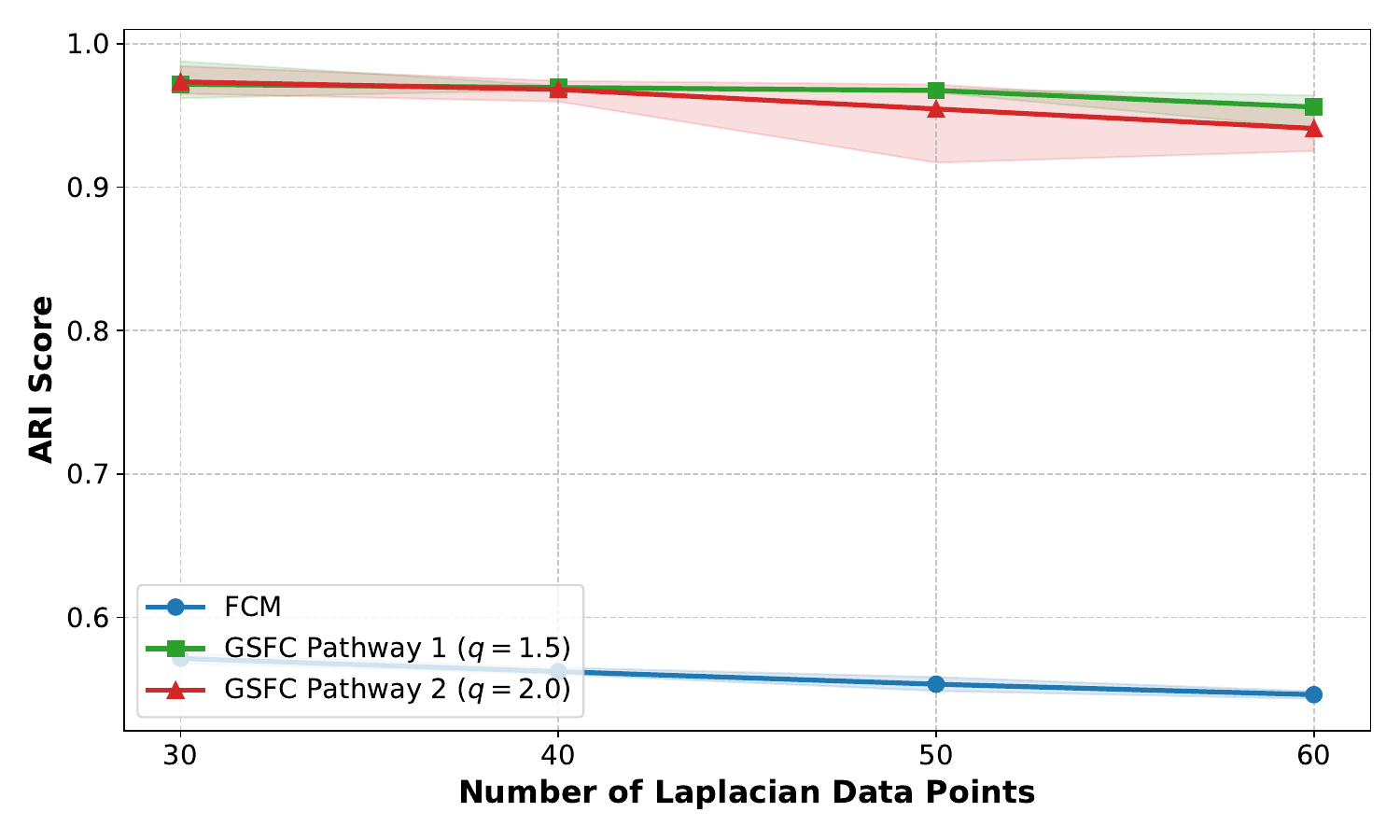}%
        }
        
        \vspace{1ex}
        \caption{Robustness evaluation of GSFC Pathways 1 and 2 compared to the FCM baseline. \textbf{(a)} ARI scores across varying levels of uniform background noise. \textbf{(b)} ARI scores as the number of  Laplacian data points increases. Solid lines denote the mean ARI across multiple initializations, while the shaded regions bound the minimum and maximum observed values.}
        \label{fig:toy_res}
    \end{minipage}
\end{center}
\subsection{Toy  Examples}
Before considering realistic   applications, we will first investigate the value of sparsity in the GSFC framework through toy $1-$dimensional experiments. More specifically, we conducted two experiments to investigate GSFC in differently shaped compact clusters: \begin{itemize}
    \item \textbf{Experiment A:}  We generated a synthetic dataset containing three clusters whose points stem from Gaussian distributions (consisting of $500$, $300$, and $150$ points, respectively) alongside uniform background noise ($N_{\rm np}$ ranging from $0$ to $300$ points) sampled from $[-12.0, 12.0]^2$. The Gaussian means are located at coordinates $[-5.0, -5.0]$, $[5.0, -5.0]$, and $[0.0, 5.0]$, with the standard deviations are equal to $2.5$ (highly diffuse), $1.2$ (moderate), and $0.4$ (highly dense). This configuration explicitly tests an algorithm's ability to adapt to non-uniform cluster densities while simultaneously resisting the effect background outliers. The number of clusters of the algorithm was initialized to $15$ and the sparsity  penalty multiplier to $s=10$. The elimination threshold was set to $15$ and the distance threshold for merging to $5$.
    \item \textbf{Experiment B:} 
    We generated a mixed-distribution dataset containing uniform background noise (sampled across a $[-8.0, 8.0]^2$ domain) and four   clusters with distinct statistical profiles: a   Gaussian cluster centered at $[-2.5, 2.0]$ (standard 0.75) with 300 points, a bounded uniform disc centered at $[-2.0, -2.5]$ (radius 2.0) with 300 points, a skewed log-normal cluster base-centered at $[0.5, -4.5]$ ($\mu=0.0$, $\sigma=0.65$) with 300 points, and a heavy-tailed Laplace cluster centered at $[2.0, 2.0]$ (scale 1.3). To explicitly test the algorithms' robustness against heavy-tailed data, the Laplace cluster is positioned so its double-exponential tail bleeds into the adjacent Gaussian cluster, and its total number of points ($N_{\rm Lap}$) is systematically varied from $30$ to $60$. For this setup, the elimination threshold was increased to $20$ and the sparsity penalty decreased to $5$ to account for more closely positioned clusters. 
\end{itemize}
These data do not have a specific geometry (e.g. line segments), and the basic GSFC algorithm with point representatives is employed. They serve merely as a preliminary testbed.

In this evaluation, we have used the Adjusted Rand Index (ARI)~\cite{ARI},  a statistical measure used in machine learning and data science to evaluate and compare the similarity of two data clusterings. Since the data are synthetic, the clustering results for each case are compared against the ground-truth grouping of the data.   All performance scores in this section constitute averages computed out of \textbf{$10$ runs} with different initialization conditions each.

\noindent \textbf{Remark:} Throughout this section, FCM and other conventional algorithms like K-means or Gaussian Mixture Models (GMM) are implemented using skfuzzy~\cite{scikit_fuzzy_2024} or sklearn~\cite{scikit-learn}. This implies that the true number of clusters is provided as an argument, hence they have a pretty significant advantage over GSFC. 

The resulting scores are reported in Fig. \ref{fig:toy_res}, where the impact of sparsity in clustering performance is evident. Fig. \ref{fig:ari_noise} reveals that FCM is highly sensitive to noise points. When there are no noise points it matches the performance of GSFC. However, as the number of such points increase its ARI score exhibits a steep decline.  Furthermore,  as shown \ref{fig:ari_laplace} the heavy-tailed Laplacian data  leads FCM to a very low ARI score compared to our method.  Both GSFC pathways accurately filter-out noise points and compute the true number of clusters, while achieving ARI scores larger than $0.9$ for all experimental configurations. Finally, as indicated by the shaded regions, GSFC displays strong robustness to initialization conditions as the number of noise/Laplacian points increases.

\subsection{UCI Wine Dataset}
To evaluate the algorithm's capability to autonomously recover cluster cardinality in higher-dimensional real-world data, we applied GSFC to the UCI Wine dataset~\cite{UCI_Dataset}. Comprising 178 samples spanning 13 standardized chemical attributes, the dataset contains three ($3$) distinct clusters. We over-initialized two GSFC configurations (Pathway 1, $q=1.5, p=0.5$; Pathway 2, $q=2.0, p=0.5$) with $m_{0} = 6$ clusters, whereas the sparsity multiplier is set to $s=0.1$ and the elimination gap to $2.0$. To provide a more comprehensive assessment of the cluster quality, we additionally evaluated the Normalized Mutual Information (NMI) and the internal cluster cohesion via the Silhouette score.

We compared the performance of GSFC against baseline FCM, K-means,  and GMMs, with the results being reported in Table \ref{tab:wine_results}. Interestingly, both pathways converge to the same exact ARI score, outperforming all benchmarks. At the same time they successfully learn the true number of clusters. This performance improvement occurs because the sparsity penalty truncates long-tail fuzzy assignments to absolute zero, preventing distant data points from skewing centroid calculations and allowing the remaining prototypes to lock perfectly onto the true density peaks.
Furthermore, a crucial observation is the  stability of GSFC under initialization conditions: across multiple independent initializations, both pathways consistently converge to identical metric scores, underscoring the algorithm's robustness against suboptimal initial cluster placements.

In Fig. \ref{fig:sensivity_wine} we evaluated both optimization pathways for different values of $s$. As is evident, the ARI scores remain consistently over $0.75$. In fact, for almost all configurations it is  higher than $0.85$,
highlighting the robustness of GSFC to the parameter $s$.  
\begin{center}
    \centering
\includegraphics[width=0.65\linewidth]{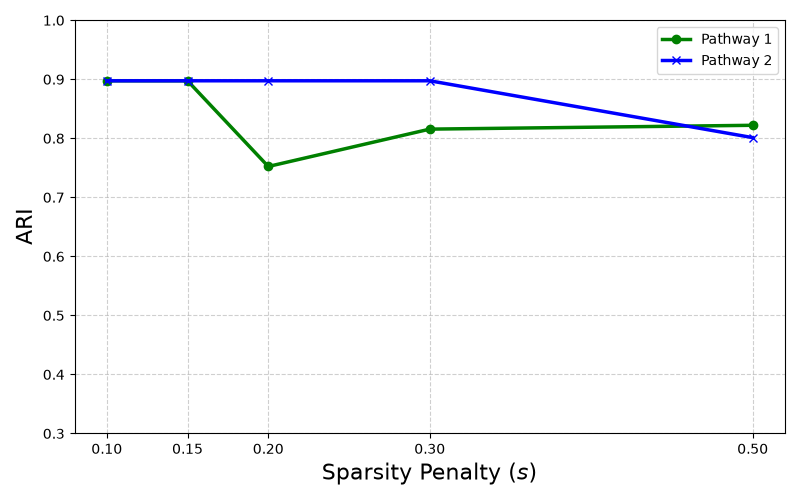}
    \captionof{figure}{\textbf{Sensitivity Study in the UCI Wine Dataset:} The average ARI scores of both pathways are reported for different values of the sparsity parameter $s$.}
    \label{fig:sensivity_wine}
\end{center}
\begin{table}[!t]
\centering
\caption{Quantitative comparison on the UCI Wine dataset across 5 independent runs. Competitors are provided with the true number of clusters ($3$), whereas GSFC autonomously recovers it from an over-initialization of $m^{(0)} = 6$.}
\label{tab:wine_results}
\begin{tabular}{llccc}
\toprule
\textbf{Metric} & \textbf{Algorithm} & \textbf{Initial $m^{(0)}$} & \textbf{Estimated $m^*$} & \textbf{Score (Mean $\pm$ Std)} \\
\midrule
\multirow{4}{*}{\textbf{ARI}} 
 & K-means & 3 & 3 & $0.8968 \pm 0.0249$ \\
 & FCM & 3 & 3 & $0.8785 \pm 0.0063$ \\
 & GMM  & 3 & 3 & $0.4084 \pm 0.0116$ \\
 & \textbf{GSFC (Pathway 1)} & 6 & \textbf{3} & $\mathbf{0.8975 \pm 0.0000}$ \\
  & \textbf{GSFC (Pathway 2)} & 6 & \textbf{3} & $\mathbf{0.8975 \pm 0.0000}$ \\
\midrule
\multirow{4}{*}{\textbf{NMI}} 
 & K-means & 3 & 3 & $0.8737 \pm 0.0245$ \\
 & FCM & 3 & 3 & $0.8464 \pm 0.0064$ \\
 & GMM & 3 & 3 & $0.5066 \pm 0.0185$ \\
 & \textbf{GSFC (Pathway 1} & 6 & \textbf{3} & $\mathbf{0.8759 \pm 0.0000}$ \\
  & \textbf{GSFC (Pathway 2} & 6 & \textbf{3} & $\mathbf{0.8759 \pm 0.0000}$ \\
\midrule
\multirow{4}{*}{\textbf{Silhouette}} 
 & K-means & 3 & 3 & $0.2597 \pm 0.0336$ \\
 & FCM & 3 & 3 & $0.2836 \pm 0.0005$ \\
 & GMM & 3 & 3 & $0.2587 \pm 0.0012$ \\
 & \textbf{GSFC (Pathway 1)} & 6 & \textbf{3} & $\mathbf{0.2849 \pm 0.0000}$ \\
  & \textbf{GSFC (Pathway 2)} & 6 & \textbf{3} & $\mathbf{0.2849 \pm 0.0000}$ \\
\bottomrule
\end{tabular}
\end{table}

%The primary objective was to determine whether the algorithm's progressive pruning logic could successfully starve redundant prototypes and converge on the true class count ($m^* = 3$) strictly through the interaction of the elimination threshold ($\epsilon_{\text{elim}}$) and the sparsity penalty ($s$), without requiring manual hyperparameter fine-tuning per run.
\subsection{Breast Cancer Wisconsin (Diagnostic) Dataset
} We considered an additional real-world tabular dataset, the Breast Cancer Wisconsin (Diagnostic) Dataset \cite{BreastCancerDataset}. It comprises of  30-dimensional features from 569 patients (i.e., 569 points), extracted from medical images. There are two official classes in the dataset, Malignant (cancerous) and Benign (non-cancerous/healthy tissue). We initialized both GSFC paradigms with $m^{(0)}=6$ clusters (3 times the number of true ground truth value). Just like the previous results, the sparsity multiplier is set to $s=0.1$ and the elimination gap to $2.0$. 

The results against FCM, K-means, and GMM are reported in Table \ref{tab:wdbc_results}. As demonstrated, GSFC Pathway 1 successfully bypasses the limitations of traditional algorithms, achieving top performance by autonomously recovering the correct cluster count ($m^* = 2$) from over-initialization while outperforming standard K-means and FCM in ARI ($0.6883$), NMI ($0.5733$), and Silhouete. In contrast, while Pathway 2 can successfully prune the cluster down to the correct number (2), its ARI and NMI are substantially smaller. Interestingly, Pathway 2 achieves the highest Silhouette score ($0.3537$) among all methods.

This paradox stems from the behavior of the quadratic penalty in high dimensions. In high-dimensional spaces, the distances between points are very large, meaning a point's distance to different centroids will not be vastly different. This leads to highly fractional memberships; for instance, most $u_{ij}$ values will be closer to $0.5$ than to $0.9$. Because these values are strictly less than 1, the quadratic term ($q=2.0$) shrinks them even more drastically ($0.9^2=0.81$ vs. $0.5^2=0.25$). To improve the cost function, the centroids are essentially "forced" apart to achieve strong exclusive memberships close to $1.0$. As these centroids are pushed toward the edges of the space, they create wide separation and thus achieve a great Silhouette score. However, the resulting decision boundary for the "middle" points is completely wrong, leading to poor performance in external metrics that use the actual labels, such as ARI and NMI.

 Again, both methods are  exceptionally robust to initialization conditions. Consider a dataset with two clusters A, B. If for instance a classical possibilistic  algorithm   initialized with 2 clusters is provided with a bad initialization with all centroids being placed inside cluster A and far away from point B, then there is a significant risk that it will get trapped to a local optimum. On the contrary, GSFC is initialized with multiple cluster representatives, meaning that each true cluster will probably have at least on or two potential representatives inside or near it.  The dense point clouds of the cluster will drag these centroids inside the cluster and close to the true center, whereas the sparsity penalty will force the membership weights of points away from them (from other clusters) to small values. When they are close to each other they will get merged into one. In contrast, potential representatives initialized away from all clusters will be pruned. Hence, GSFC mitigates the initial noise due to the centroid initialization, displaying profound robustness across all datasets considered.
\begin{table}[!t]
\centering
\caption{Quantitative comparison on the Breast Cancer Wisconsin (Diagnostic) (WDBC) dataset across 5 independent runs. Competitors are provided with the true number of clusters ($2$), whereas GSFC autonomously recovers or evaluates it from an over-initialization of $m^{(0)} = 6$.}
\label{tab:wdbc_results}
\begin{tabular}{llccc}
\toprule
\textbf{Metric} & \textbf{Algorithm} & \textbf{Initial $m^{(0)}$} & \textbf{Estimated $m^*$} & \textbf{Score (Mean $\pm$ Std)} \\
\midrule
\multirow{5}{*}{\textbf{ARI}} 
& K-means & 2 & 2 & $0.6662 \pm 0.0105$ \\
& FCM & 2 & 2 & $0.6829 \pm 0.0000$ \\
& GMM & 2 & 2 & $0.0000 \pm 0.0000$ \\
& \textbf{GSFC (Pathway 1)} & 6 & \textbf{2} & $\mathbf{0.6883 \pm 0.0000}$ \\
& \textbf{GSFC (Pathway 2)} & 6 & \textbf{2} & $0.4245 \pm 0.0000$ \\
\midrule
\multirow{5}{*}{\textbf{NMI}} 
& K-means & 2 & 2 & $0.5481 \pm 0.0134$ \\
& FCM & 2 & 2 & $0.5612 \pm 0.0000$ \\
& GMM & 2 & 2 & $0.0000 \pm 0.0000$ \\
& \textbf{GSFC (Pathway 1)} & 6 & \textbf{2} & $\mathbf{0.5733 \pm 0.0000}$ \\
& \textbf{GSFC (Pathway 2)} & 6 & \textbf{2} & $0.3614 \pm 0.0000$ \\
\midrule
\multirow{5}{*}{\textbf{Silhouette}} 
& K-means & 2 & 2 & $0.3432 \pm 0.0009$ \\
& FCM & 2 & 2 & $0.3400 \pm 0.0000$ \\
& GMM & 2 & 2 & $0.0000 \pm 0.0000$ \\
& \textbf{GSFC (Pathway 1)} & 6 & \textbf{2} & $0.3433 \pm 0.0000$ \\
& \textbf{GSFC (Pathway 2)} & 6 & \textbf{2} & $\mathbf{0.3537\pm 0.0000}$ \\
\bottomrule
\end{tabular}
\end{table}
\color{black}
\subsection{HyperSpectral Images (HSIs)}
We will now evaluate GSFC on three publicly available HSIs. The Salinas-A scene~\cite{salinas_url}, the Pavia university scene~\cite{pavia_url}, and the AVIRIS Cuprite dataset~\cite{aviris_url}. All three datasets contain ground truth labels for the pixels and hence we can evaluate GSFC using metrics like the ARI score.
The following pre-processing pipeline was performed for all three images:
\begin{enumerate}
    \item Pixels with a zero-class label are excluded. % to restrict evaluation strictly to homogeneous agricultural regions.
\item Spectral signatures are normalized to zero mean and unit variance.
\item Finally,  PCA projects the $200+$ bands down to $d=10$ components, mitigating high-frequency noise while preserving spatial variance.
\end{enumerate}

The Salinas-A hyperspectral scene, captures an agricultural area in Salinas Valley, California. The image comprises a $220 \times 120$ spatial grid with a spatial resolution of $3.7$ meters per pixel, yielding a total of $N = 26,400$ sample pixels across $204$ spectral bands. The ground truth defines seven distinct crop classes (grapes, broccoli, fallow 1, fallow 2, fallow 3, stubble, and celery), while unlabeled background pixels are masked out from the evaluation. 
To demonstrate autonomous structural recovery, GSFC  is heavily over-initialized with $15$ candidate prototypes. The sparsity parameter is set to $s = 0.01$.  In
highly mixed datasets (like mineral mixtures or diffuse distributions), a larger $s$ may be required. However, agricultural hyperspectral scenes like Salinas-A consist of human-planted blocks where crop types change abruptly at field boundaries rather than blending smoothly.
The cardinality pruning threshold is set to $\epsilon_{elim} = 500.0$  (requiring a cluster to claim roughly $2\%$ of the image's total membership weight to justify its existence) and the coalescence gap to $\epsilon_{gap} = 5.0$.

For reference, we  again considered the classic K-means algorithm, the FCM and the GMM (again the  implementations require precise specification of the number of clusters). GSFC correctly infers the true number of clusters, and the ARI   scores presented in Fig. \ref{fig:SalinasRes} illustrate that it outperforms all baselines. Interestingly, classic FCM is highly sensitive to initialization conditions. For some "lucky" runs it comes close to  matching the ARI scores of GSFC but for some other "unlucky" runs it outputs very low-quality clusterings.
%\begin{figure}[!tb]
%    \centering
%\includegraphics[width=0.35\linewidth]{HFRI-Salinas-Res.png}
%    \caption{ARI scores for the Salinas-A HSI task. }
%    \label{fig:SalinasRes}
%\end{figure}
\begin{center}
    \includegraphics[width=0.65\linewidth]{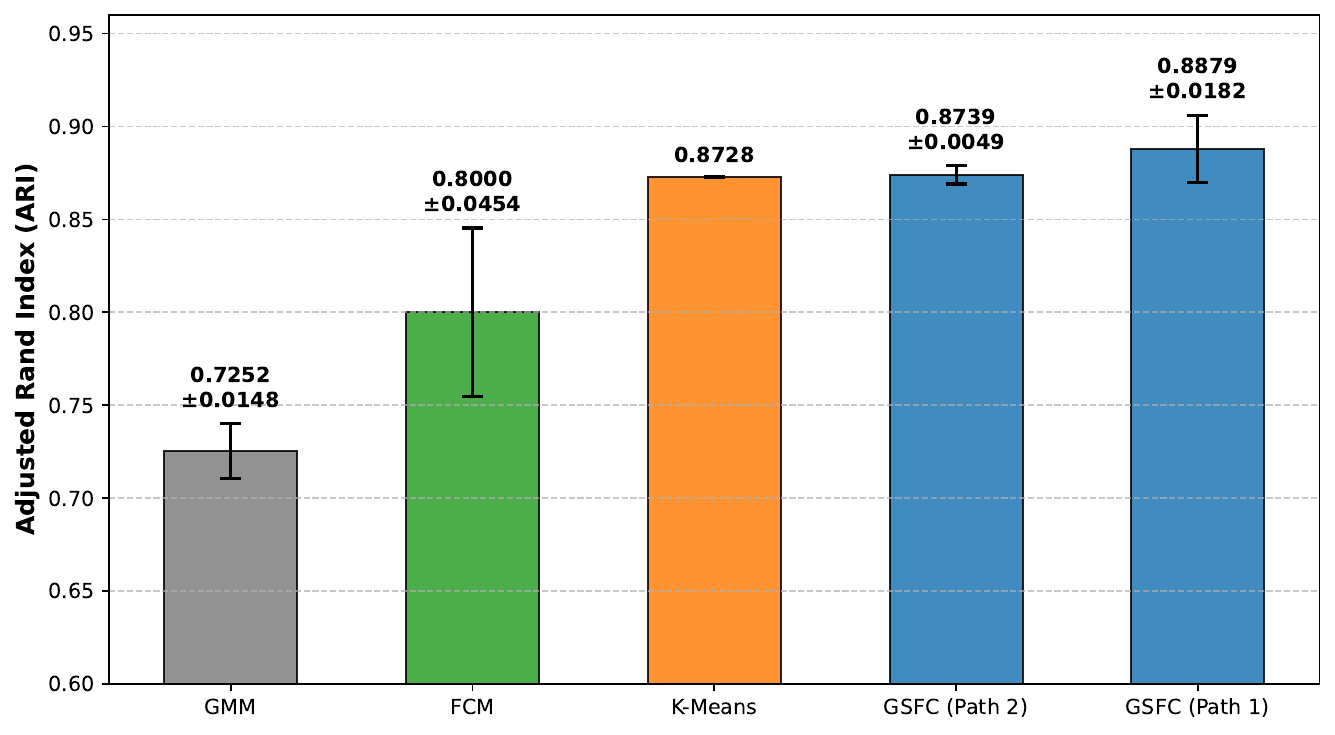}
    \captionof{figure}{ARI scores for the Salinas-A HSI task.}
    \label{fig:SalinasRes}
\end{center}

While Salinas-A contains a rural scene, in order to have a broad dataset pool, we also considered the Pavia university  image , which captures a complex urban campus environment in Pavia, northern Italy.  The HSI comprises $610 \times 340$ pixels across 103 usable spectral bands. Excluding the unlabeled background, the dataset contains 42,776 valid pixels distributed across 9 distinct ground-truth classes, representing diverse materials. As shown in Fig. \ref{fig:PaviaRes} the GSFC (with the same parameter configuration as the one used for the Salinas-A image) outperforms all baselines in terms of ARI by a significant margins, while also outputing the correct number of clusters. Its dense counterpart performs extremely poorly.

\begin{center}
    \includegraphics[width=0.65\linewidth]{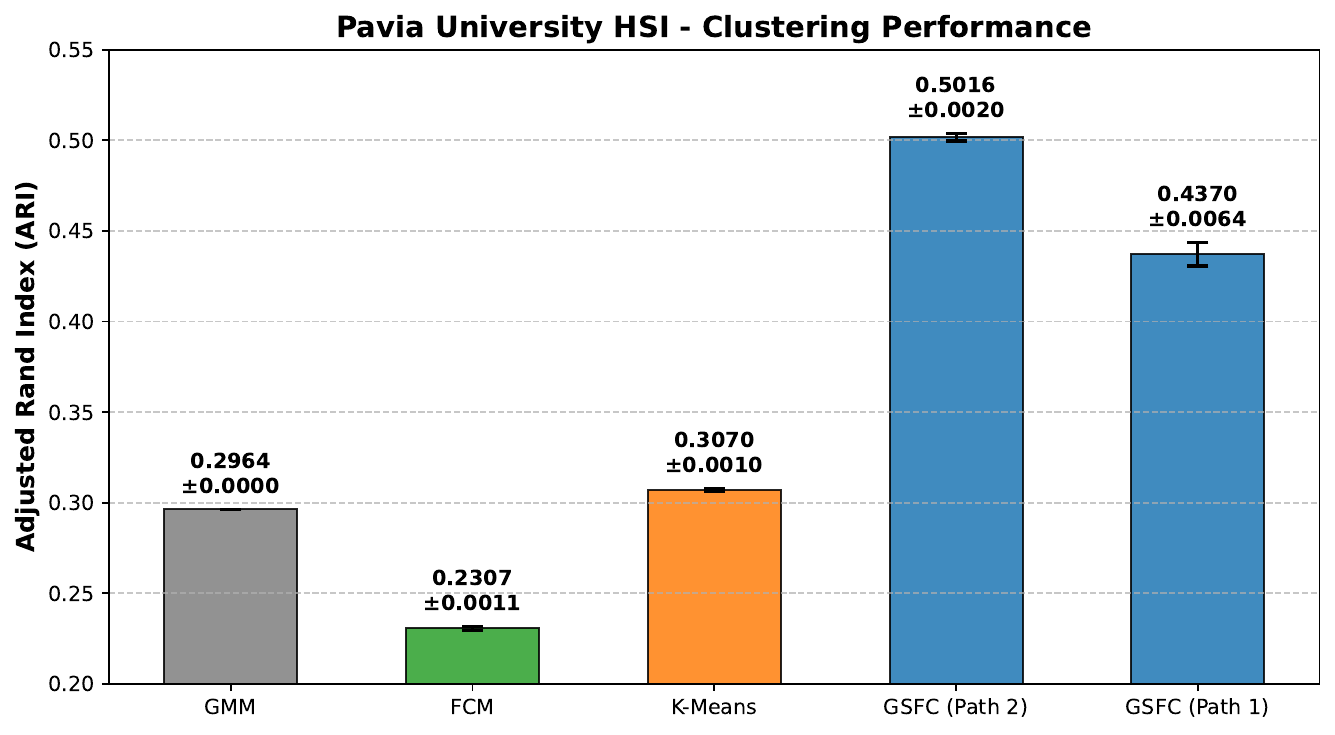}
    \captionof{figure}{ARI scores for the Pavia university HSI task.}
    \label{fig:PaviaRes}
\end{center}

Finally, the AVIRIS Cuprite HSI, captured by NASA's Airborne Visible Infrared Imaging Spectrometer over the Cuprite Hills mining district in Nevada, serves as a benchmark for geological and mineralogical remote sensing. The scene is globally renowned for its exposed, diverse mineral types (such as alunite, buddingtonite, calcite, and kaolinite). Unlike the neatly managed agricultural blocks of Salinas-A, the Cuprite scene is characterized by complex, irregular mineral outcroppings interspersed with extensive background scatter, atmospheric noise, and pervasive mixed-pixel boundaries. These geological characteristics create a challenging environment where traditional clustering algorithms frequently fail due to their susceptibility to background interference.

While the Cuprite scene encompasses a broad range of potential mineral types across the entire landscape, any individual pixel physically comprises only a small subset of those elements. This localized reality provides an ideal physical justification for GSFC's sparsity component: rather than forcing a pixel's membership to leak across numerous irrelevant mineral classes (as standard dense fuzzy clustering tends to do), the sparsity penalty mathematically enforces that a pixel's composition is active in only a few matching clusters.

The ARI scores are depicted in Fig. \ref{fig:AVIRIS-RESULTS}. To evaluate the robustness/sensitivity to the choice of $s$ we actually sweeped $s$ from a very small value of $0.01$ to $10$ (the rest of the parameters are set to the same values as the previous HSI experiments).  The evaluation on the Cuprite dataset demonstrates that both GSFC pathways consistently outperform classical baselines like K-means, GMM, and standard FCM across a wide range of sparsity configurations (despite also needing to learn the true number of clusters). Furthermore, the performance exhibits strong empirical robustness to the choice of $s$; except for Pathway 2 under excessively large sparsity penalties where performance degrades, the resulting ARI scores remain superior to all competing benchmarks, confirming the stability and efficacy of the noise-aware sparse framework.  Pathway 1 correctly infers $m^*$ for all considered values of $s$, and Pathway 2 for all $s \leq 2$. If  $s>2$ it slightly over-prunes estimating one less cluster.  

\begin{center}
    \includegraphics[width=0.65\linewidth]{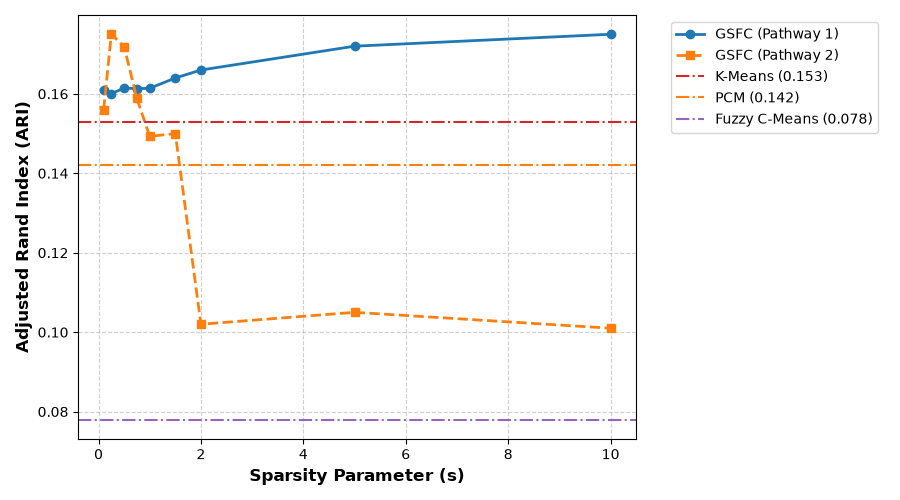}
    \captionof{figure}{ARI scores for the AVIRIS Cuprite HSI task.}
    \label{fig:AVIRIS-RESULTS}
\end{center}

\subsection{Line Segment Identification}
\begin{figure}[!tb]
\begin{subfigure}[b]{0.235\textwidth}
        \centering
        \includegraphics[width=\textwidth]{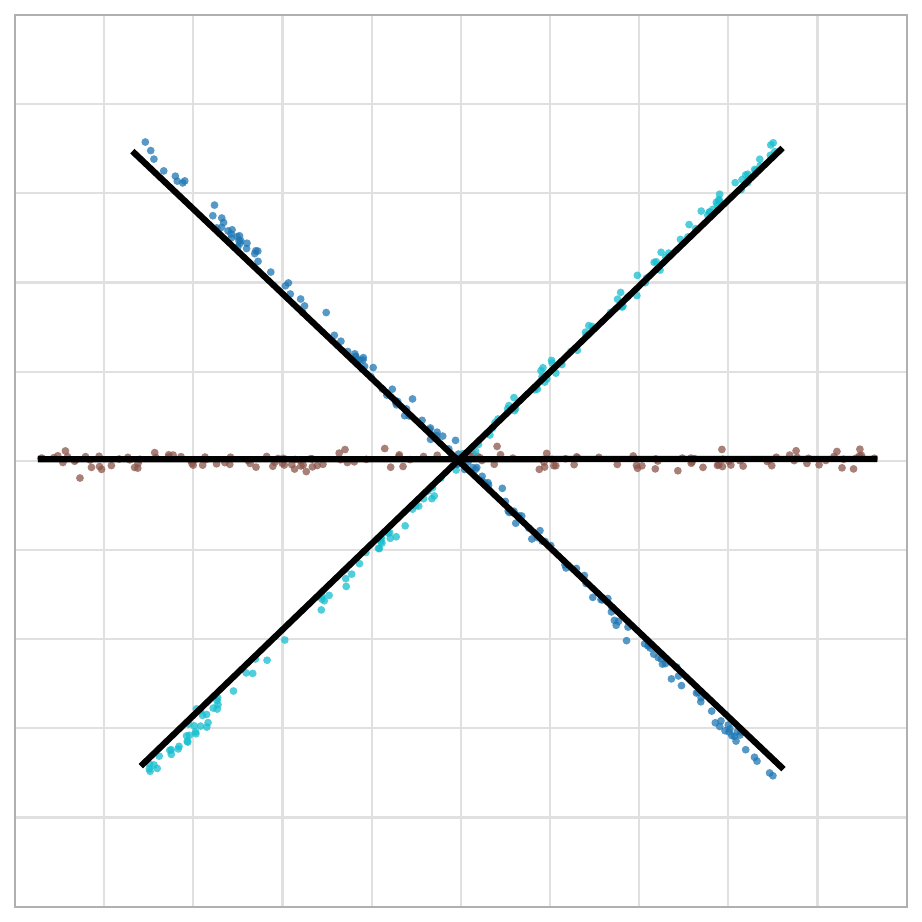}
        \caption{SPCLS \cite{K18}}
        \label{fig:case1_k18}
    \end{subfigure}
    \hfill
    \begin{subfigure}[b]{0.235\textwidth}
        \centering
        \includegraphics[width=\textwidth]{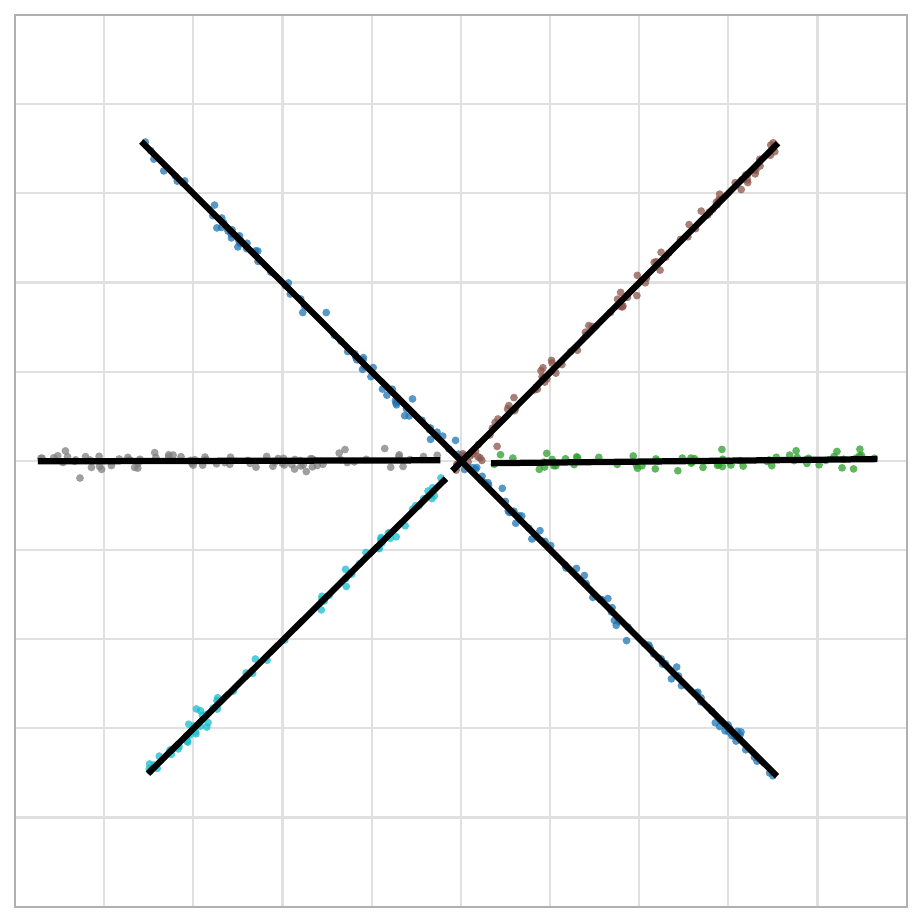}
        \caption{PLSC  \cite{S_and_K21}}
        \label{fig:case1_sk21}
    \end{subfigure}
    \hfill
    \begin{subfigure}[b]{0.235\textwidth}
        \centering
        \includegraphics[width=\textwidth]{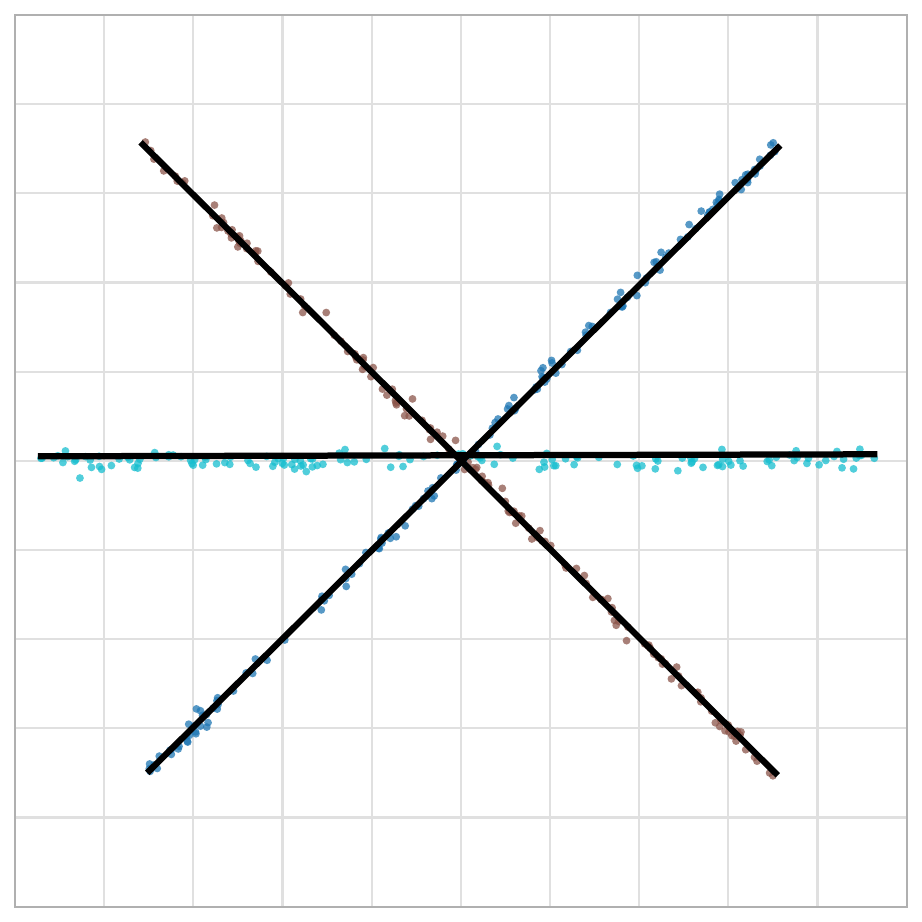}
        \caption{GSFC Pathway 1}
        \label{fig:case1_p1}
    \end{subfigure}
    \hfill
    \begin{subfigure}[b]{0.235\textwidth}
        \centering
        \includegraphics[width=\textwidth]{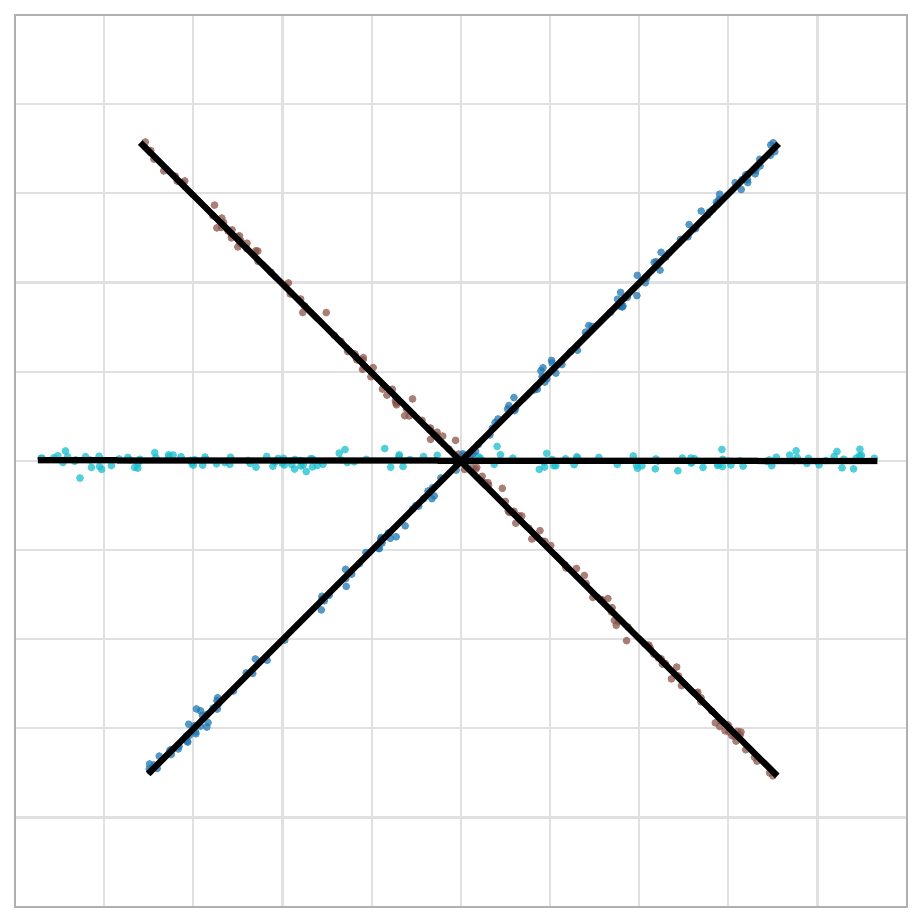}
        \caption{GSFC Pathway 2}
        \label{fig:case1_p2}
    \end{subfigure}
    \caption{Case 1: \textbf{3 Line segments intersecting at a single point.} While all methods can reconstruct the underlying geometry, the method of \cite{S_and_K21} introduces slight over-segmentation overestimating the true number of clusters. Both pathways and the algorithm of \cite{K18} estimate the correct number of clusters and produce the correct line segments.}
    \label{fig:case1_results}
\end{figure}

\begin{figure}[!tb]
\centering
\begin{subfigure}[b]{0.235\textwidth}
    \centering
    \includegraphics[width=\linewidth]{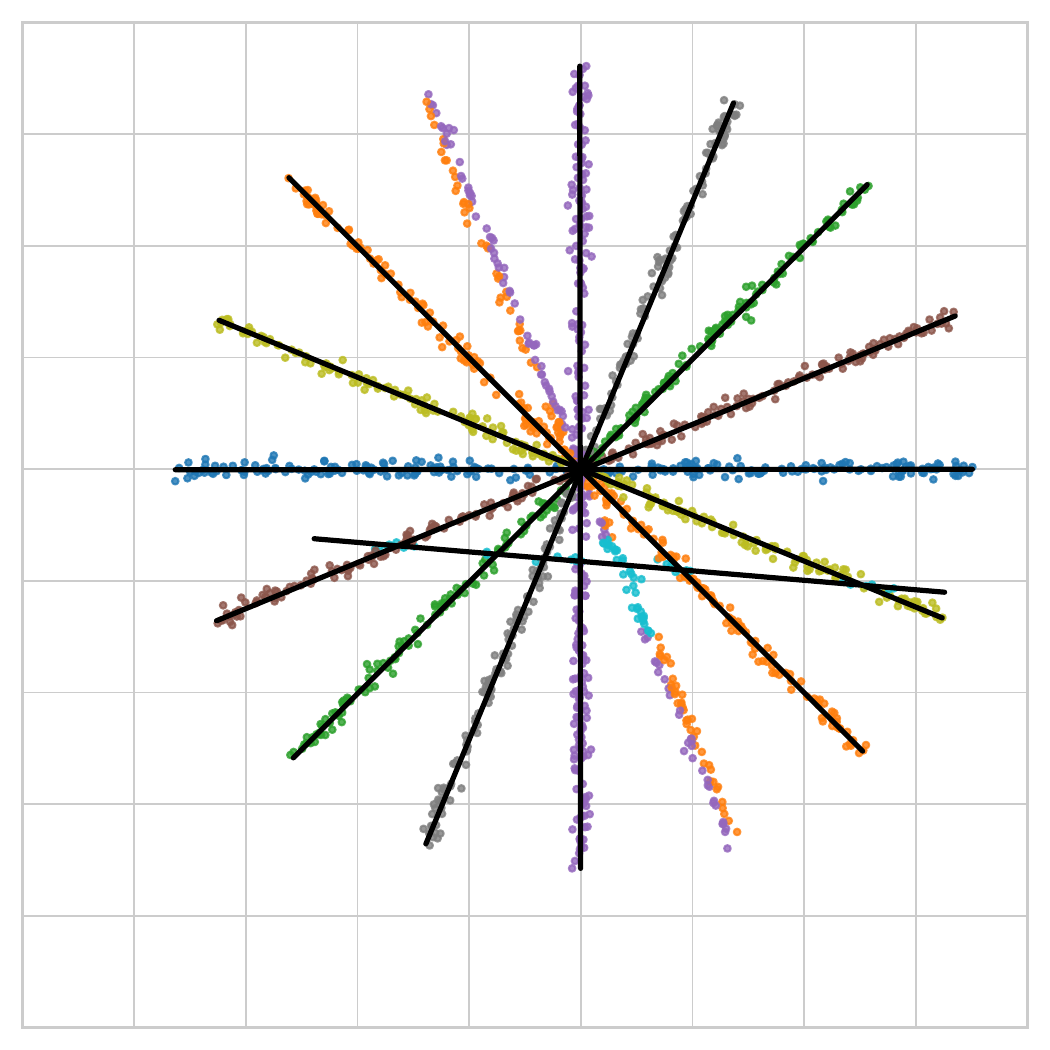}
    \caption{SPCLS~\cite{K18}}
    \label{fig:starburst_k18}
\end{subfigure}
\begin{subfigure}[b]{0.235\textwidth}
    \centering
    \includegraphics[width=\linewidth]{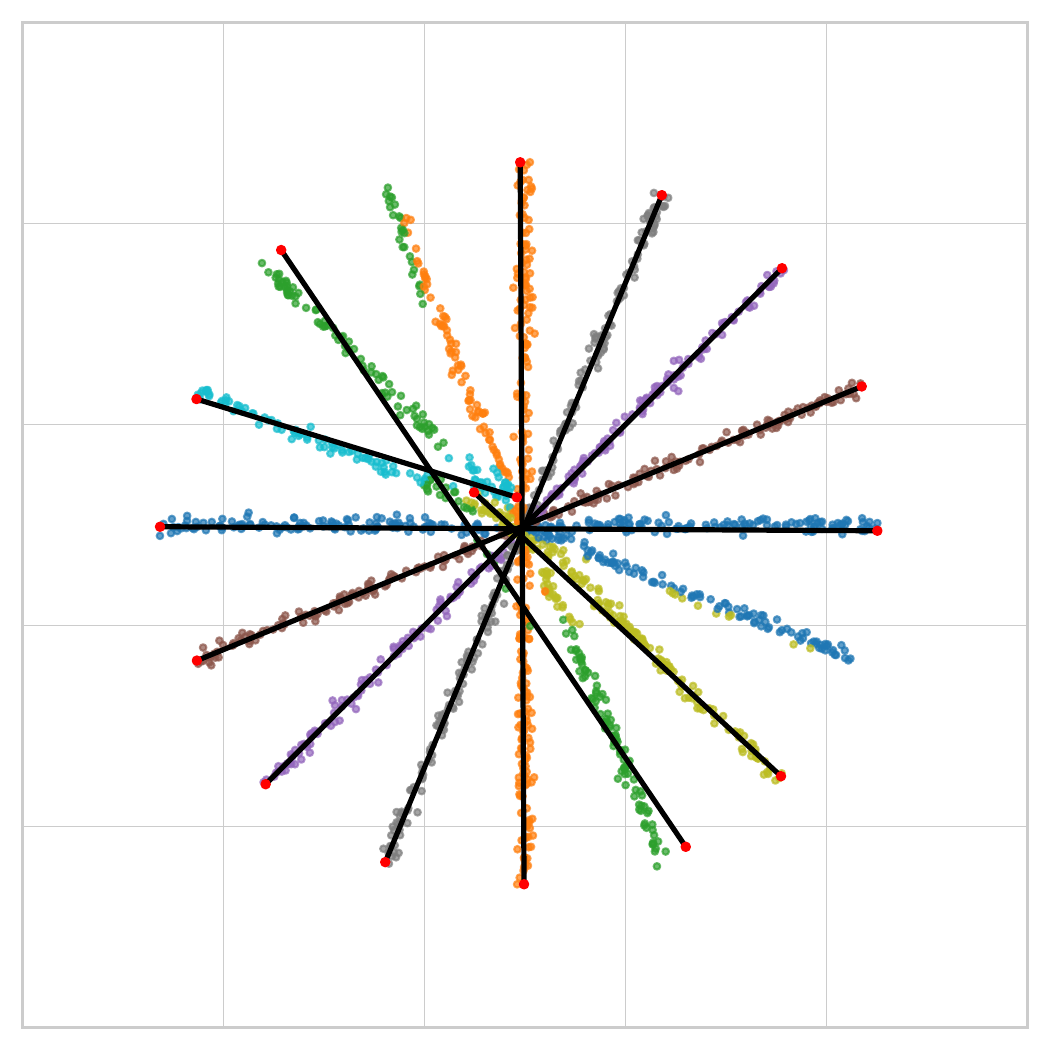}
    \caption{PLSC~\cite{S_and_K21}}
    \label{fig:starburst_sk21}
\end{subfigure}\hfill
\begin{subfigure}[b]{0.235\textwidth}
    \centering
    \includegraphics[width=\linewidth]{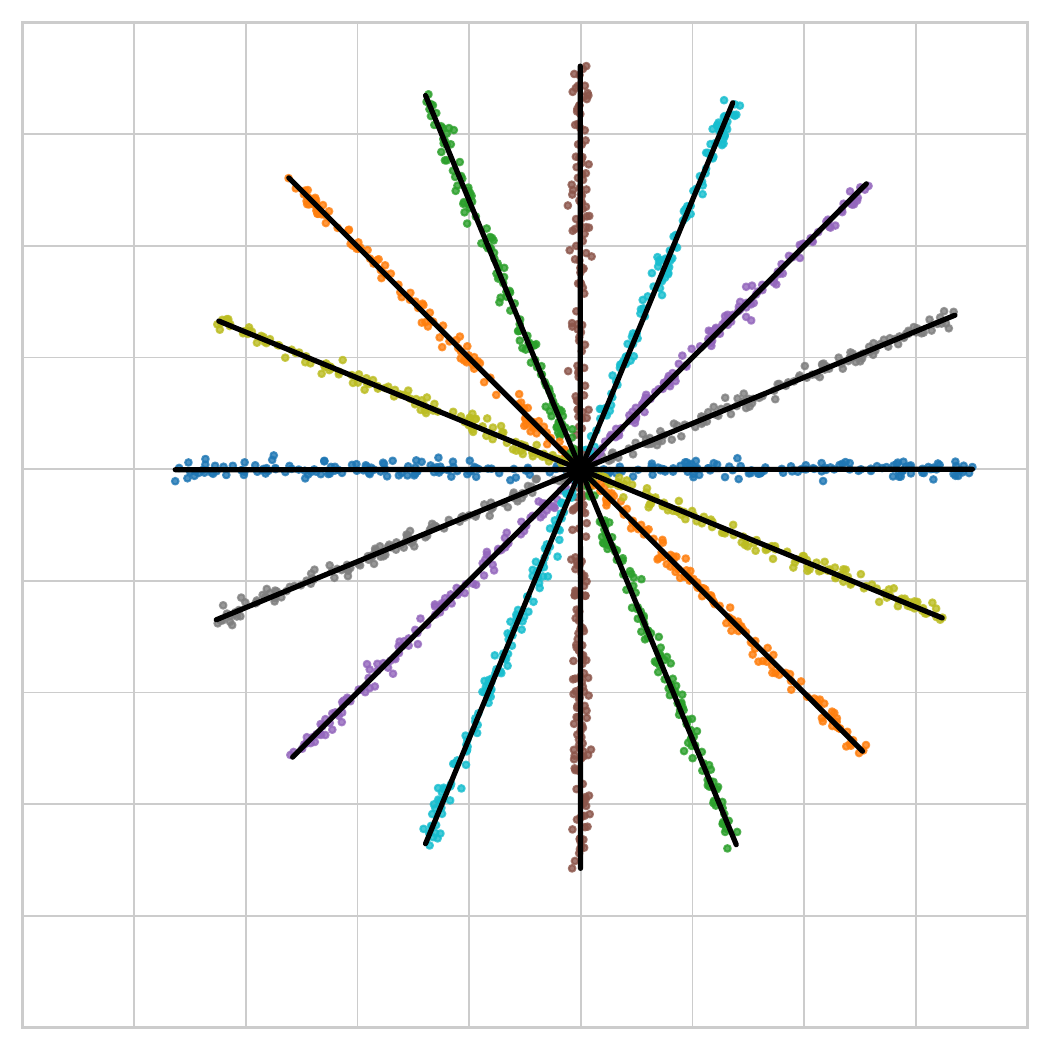}
    \caption{GSFC Pathway 1}
    \label{fig:starburst_gsfc1}
\end{subfigure}\hfill
\begin{subfigure}[b]{0.235\textwidth}
    \centering
    \includegraphics[width=\linewidth]{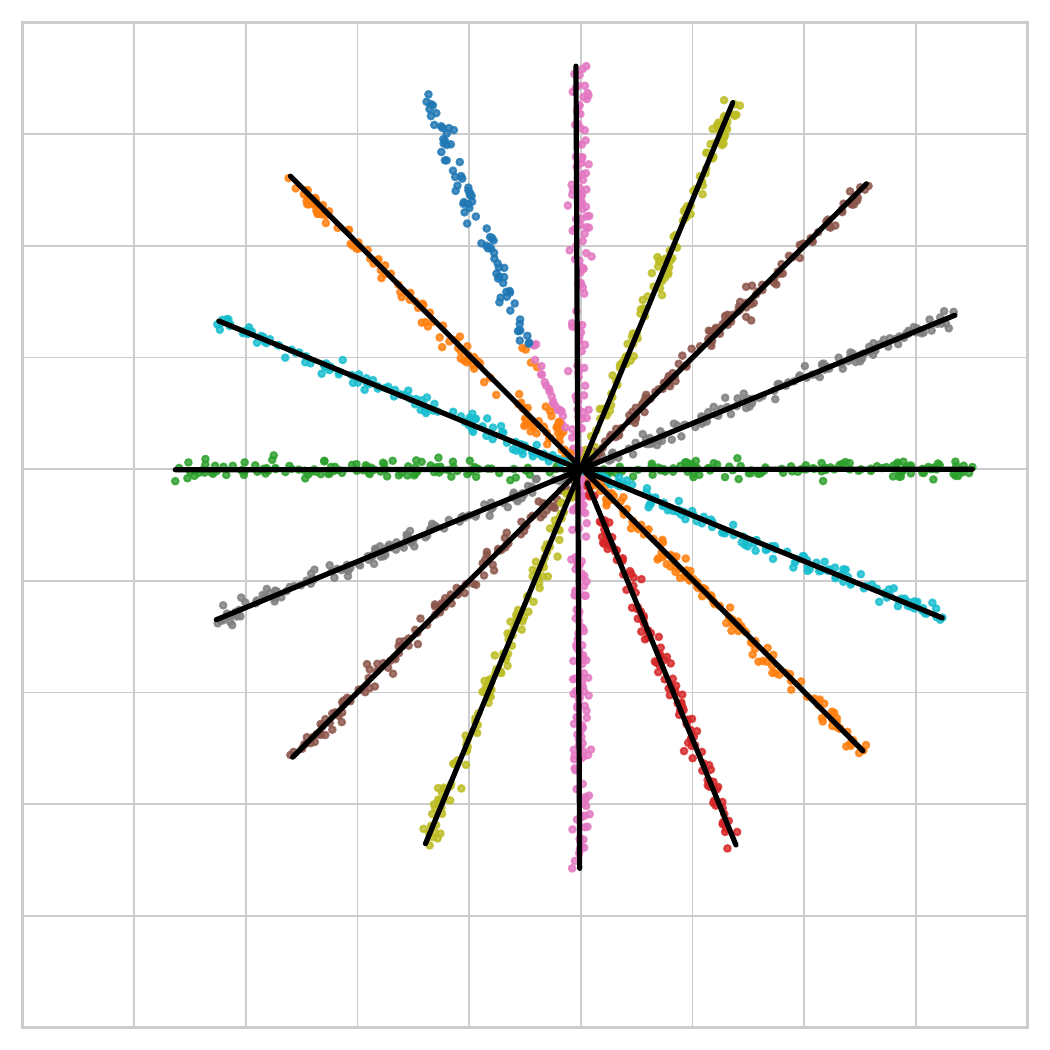}
    \caption{GSFC Pathway 2}
    \label{fig:starburst_gsfc2}
\end{subfigure}\hfill
\caption{ Case 2: \textbf{8 Line segments intersecting at a single point.}
The algorithm of \cite{S_and_K21} suffers from hub fracturing: hard Bayes partitioning creates spatial contention at the dense junction, leaving several components prematurely truncated as radial half-spokes, while \cite{K18} exhibits phantom secants: unconstrained possibilistic memberships accumulate mass transversely across adjacent rays, yielding spurious chords near the hub while dropping one true diameter. 
Finally, the proposed GSFC can accurately infer the true geomery for both optimization pathways.}
\label{fig:starburst_comparison}
\end{figure}

\begin{figure}[!tb]
\centering
\begin{subfigure}[b]{0.235\textwidth}
    \centering
    \includegraphics[width=\linewidth]{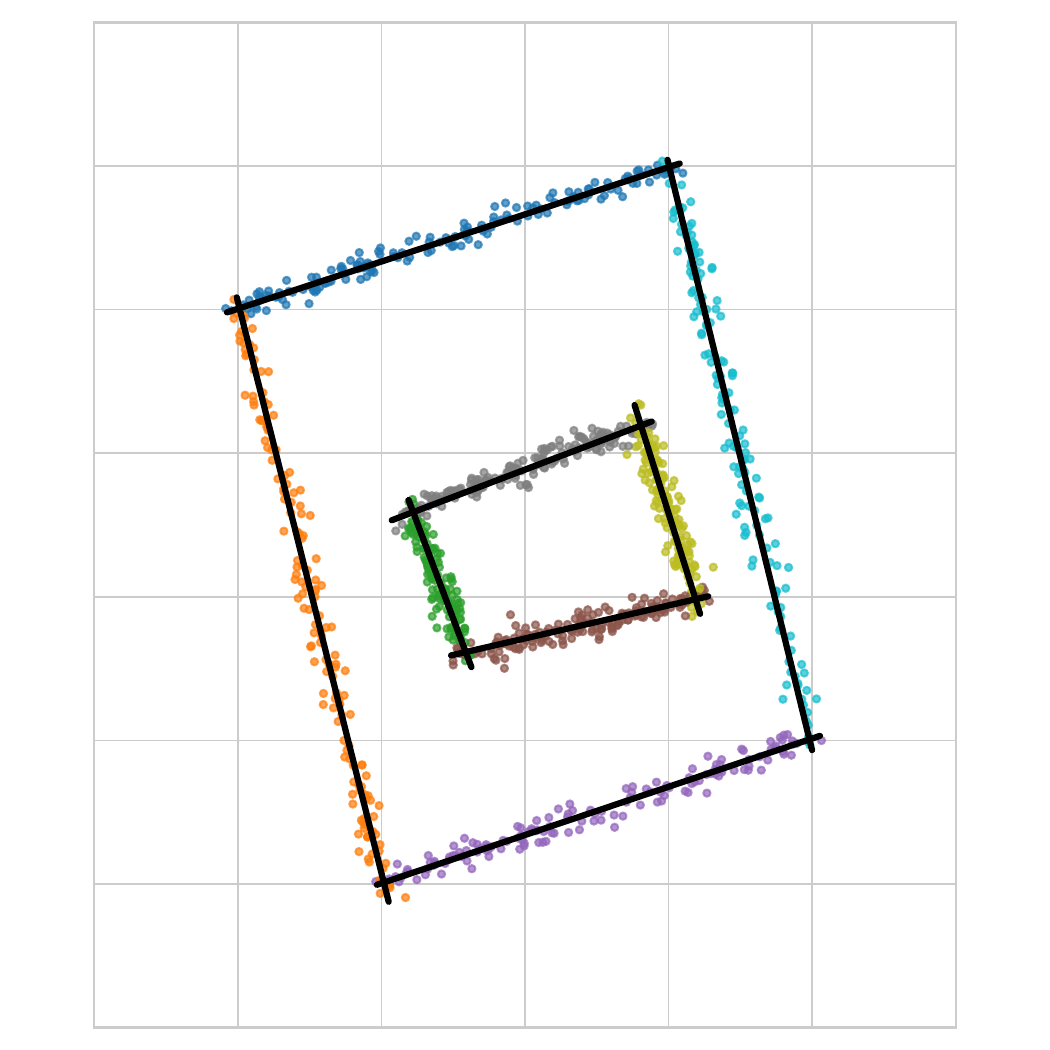}
    \caption{SPCLS~\cite{K18}}
\end{subfigure}\hfill
\begin{subfigure}[b]{0.235\textwidth}
    \centering
    \includegraphics[width=\linewidth]{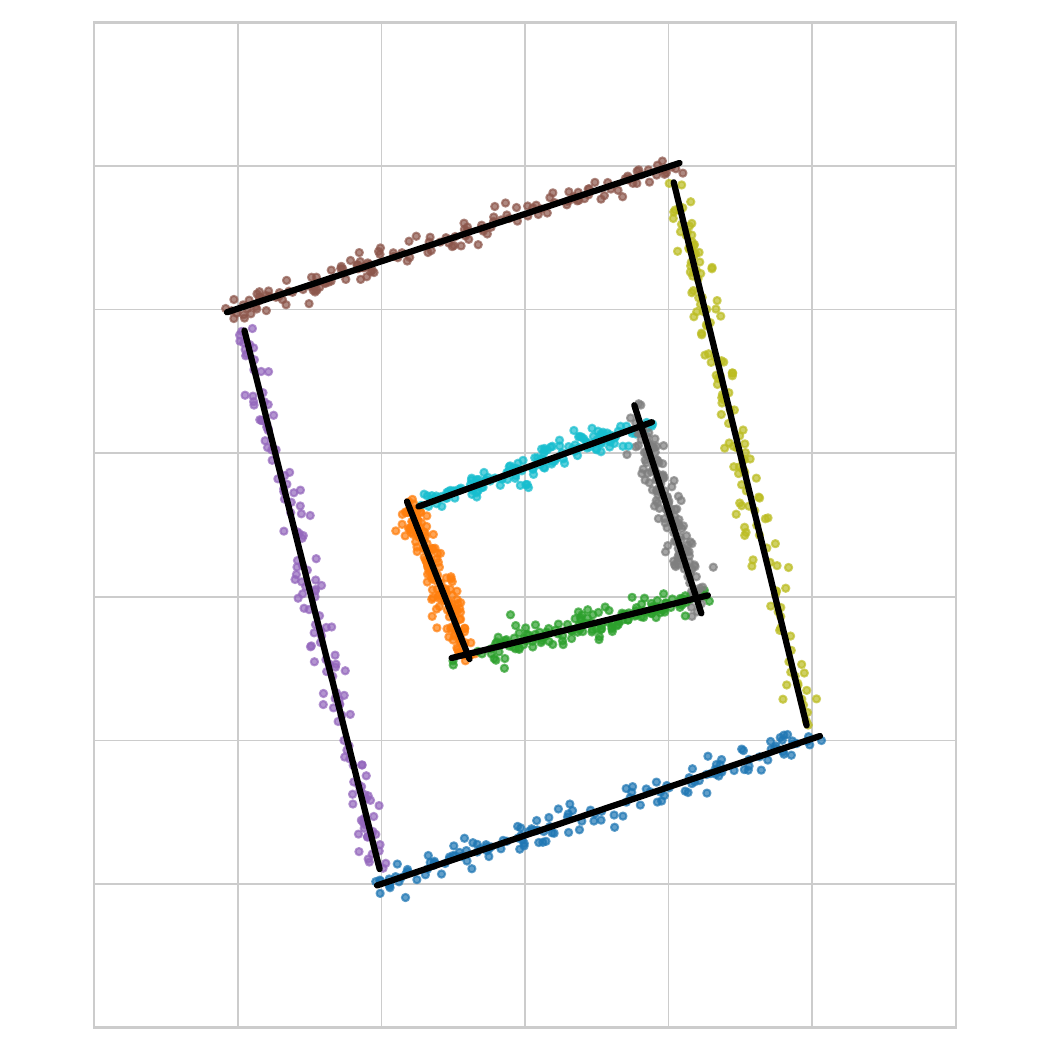}
    \caption{PLSC~\cite{S_and_K21}}
\end{subfigure}\hfill
\begin{subfigure}[b]{0.235\textwidth}
    \centering
    \includegraphics[width=\linewidth]{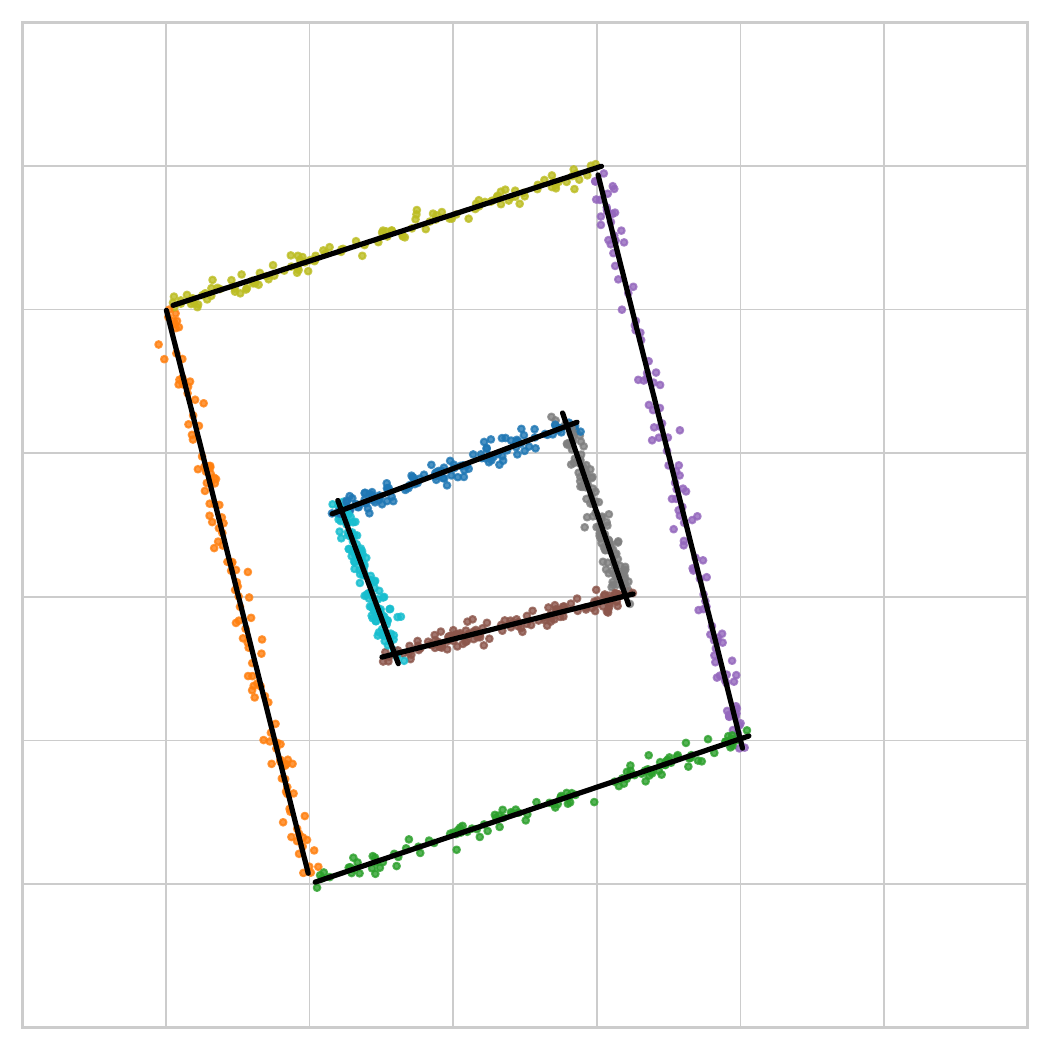}
    \caption{GSFC Pathway 1}
\end{subfigure}\hfill
\begin{subfigure}[b]{0.235\textwidth}
    \centering
    \includegraphics[width=\linewidth]{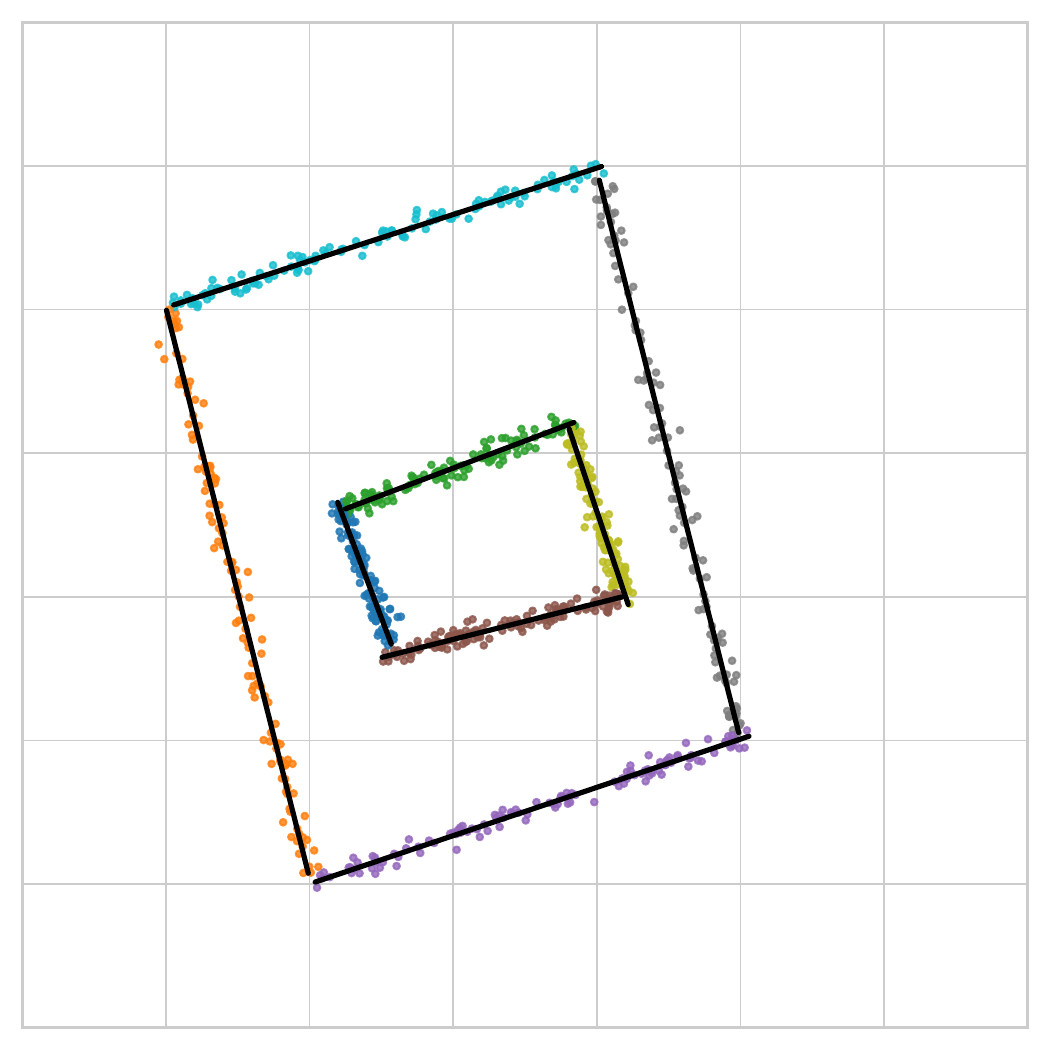}
    \caption{GSFC Pathway 2}
\end{subfigure}
\caption{ Case 3: \textbf{Overlapping Rectangles/Floppy Disk.} All algorithms can learn the underlying segments' position and orientation, with minor performance difference. The method of \cite{S_and_K21} results in slightly shorter outside vertical segments (small empty spaces).
}
\label{fig:floppy_disk}
\end{figure}

\begin{figure}[!tb]
    \centering
    \begin{subfigure}[b]{0.235\textwidth}
        \centering
        \includegraphics[width=\textwidth]{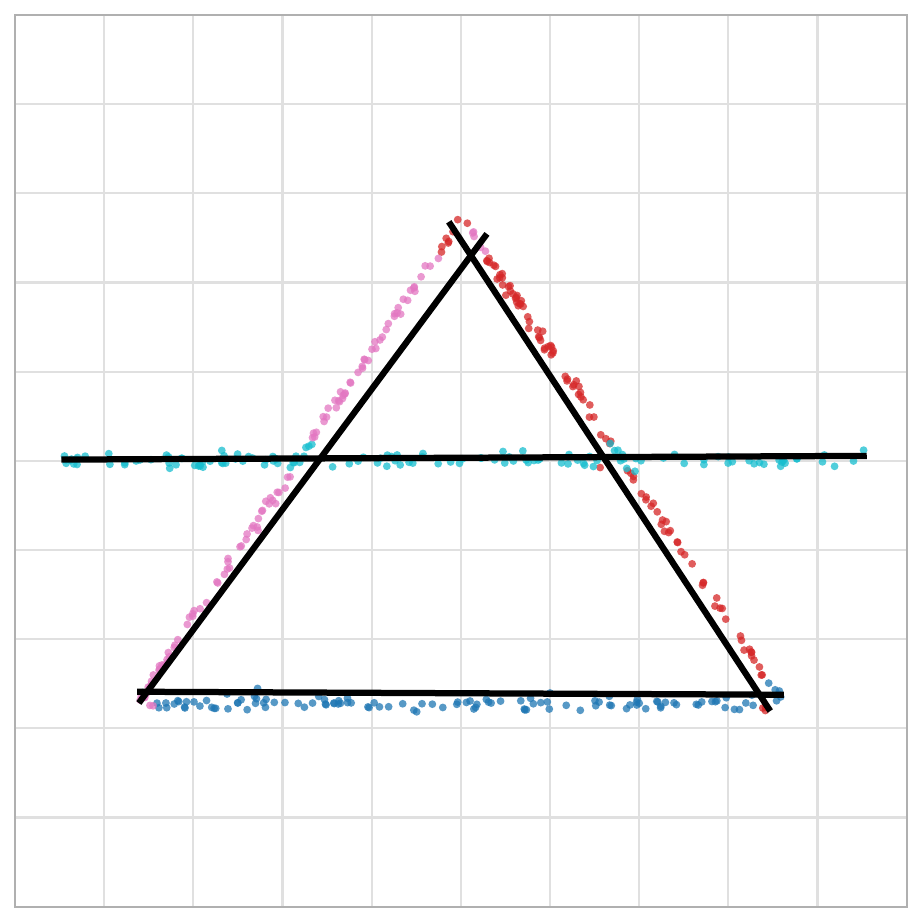}
        \caption{SPCLS~\cite{K18}}
        \label{fig:case2_k18}
    \end{subfigure} \hfill
    \begin{subfigure}[b]{0.235\textwidth}
        \centering
        \includegraphics[width=\textwidth]{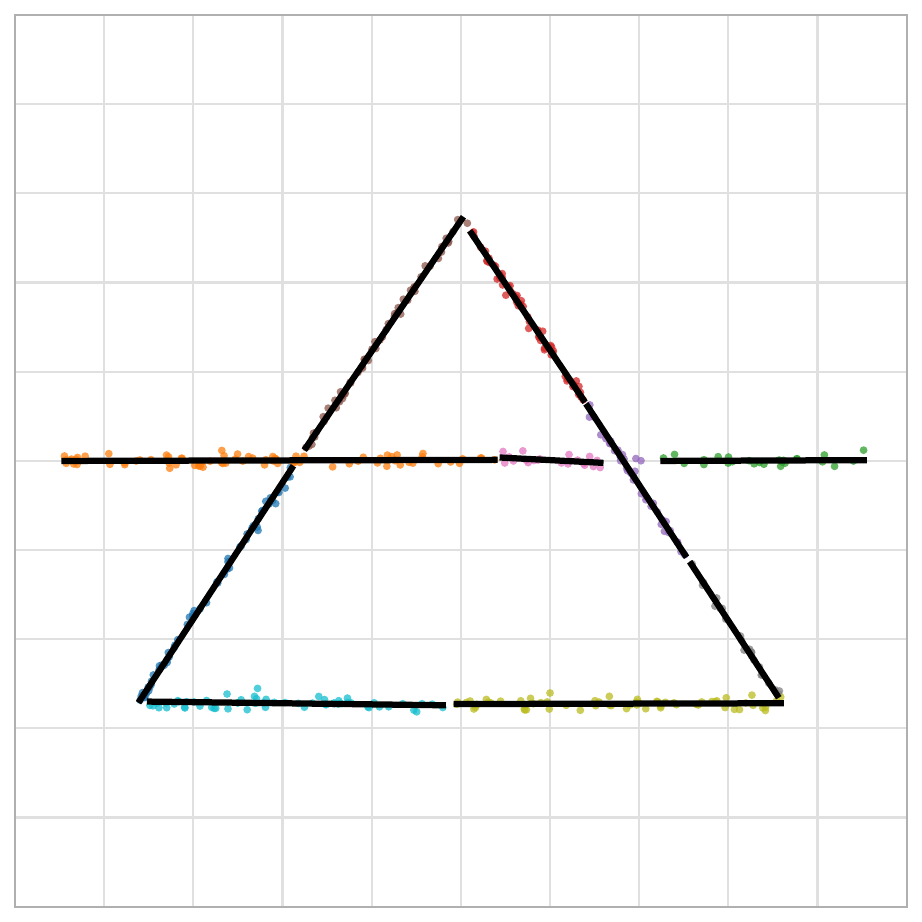}
        \caption{PLSC~\cite{S_and_K21}}
        \label{fig:case2_sk21}
    \end{subfigure}
    \hfill
    \begin{subfigure}[b]{0.235\textwidth}
        \centering
        \includegraphics[width=\textwidth]{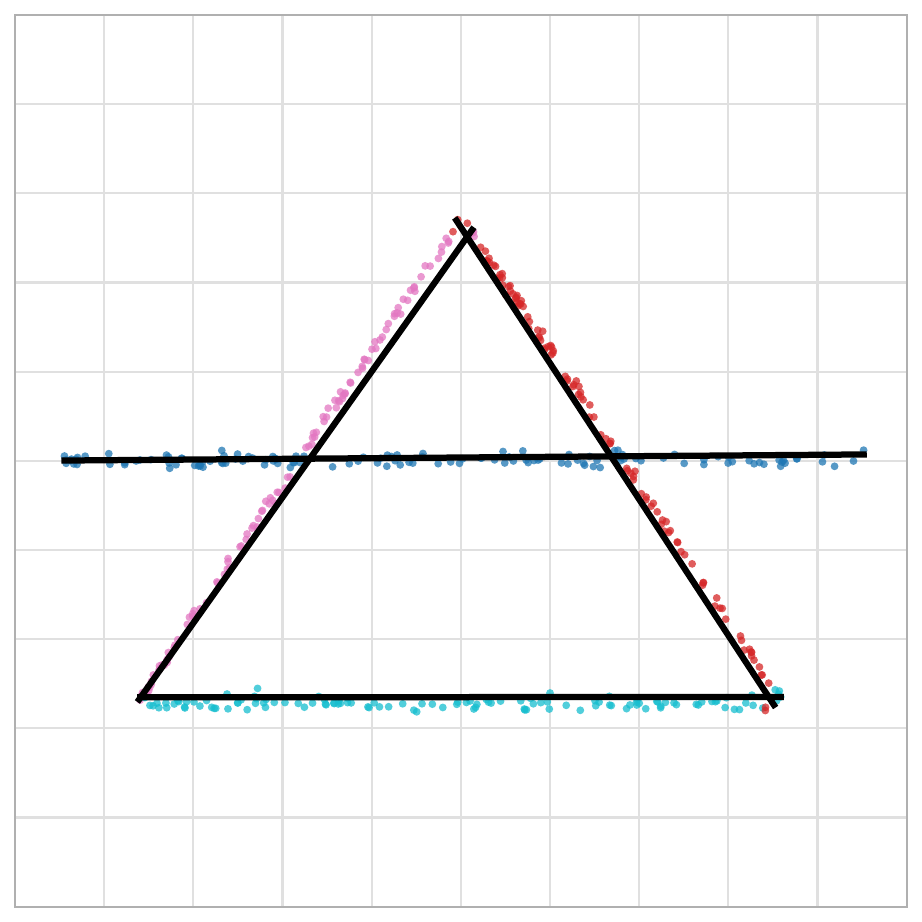}
        \caption{GSFC Pathway 1}
        \label{fig:case2_p1}
    \end{subfigure}
    \hfill
    \begin{subfigure}[b]{0.235\textwidth}
        \centering
        \includegraphics[width=\textwidth]{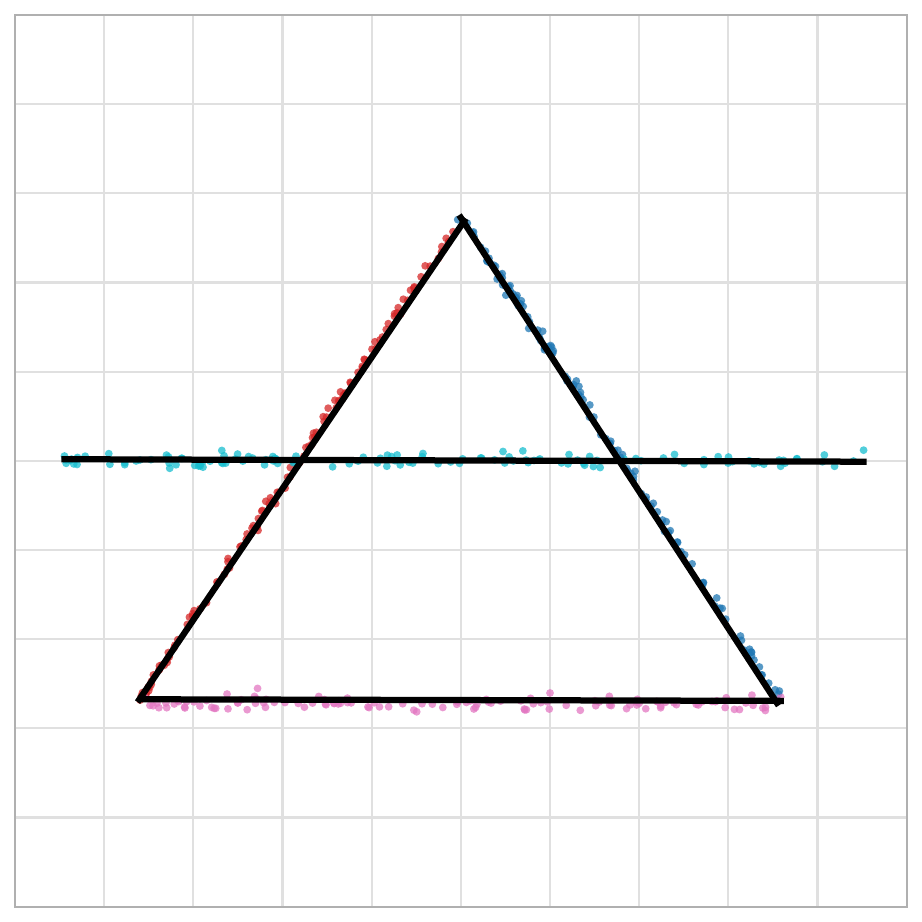}
        \caption{GSFC Pathway 2}
        \label{fig:case2_p2}
    \end{subfigure}
    \caption{Case 4: \textbf{Triangle with Horizonal Intersection}  Both proposed GSFC pathways accurately reconstruct the complex intersecting geometry of the four distinct line segments  without fragmentation (although the left triangle side is slightly  twisted for pathway 1). The method of \cite{S_and_K21} severely over-segments the structure resulting in 10 fragmented clusters. While the method of \cite{K18} identifies the correct number of clusters, the intersections act as heavy 'anchors', physically twisting the estimated segments away from their ground-truth values.}
    \label{fig:case2_results}
\end{figure}

\begin{figure}[!tb]
    \centering
    \begin{subfigure}[b]{0.235\textwidth}
        \centering
        \includegraphics[width=\textwidth]{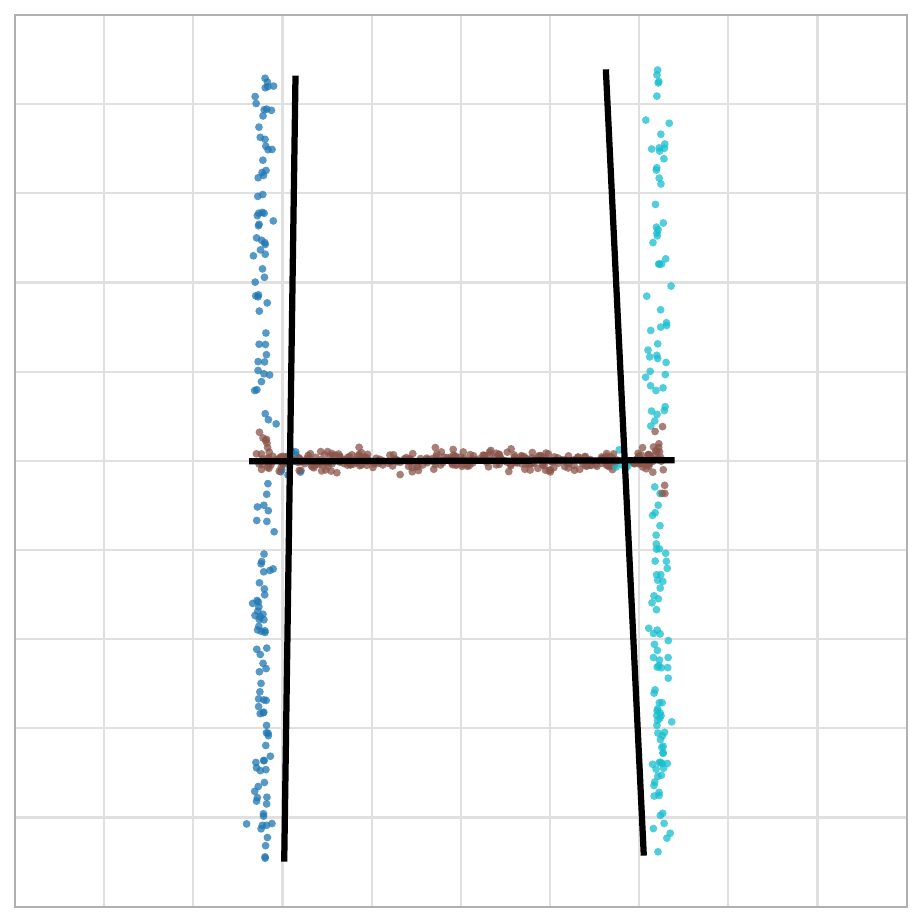}
        \caption{SPCLS~\cite{K18}}
        \label{fig:case3_k18}
    \end{subfigure}
    \begin{subfigure}[b]{0.235\textwidth}
        \centering
        \includegraphics[width=\textwidth]{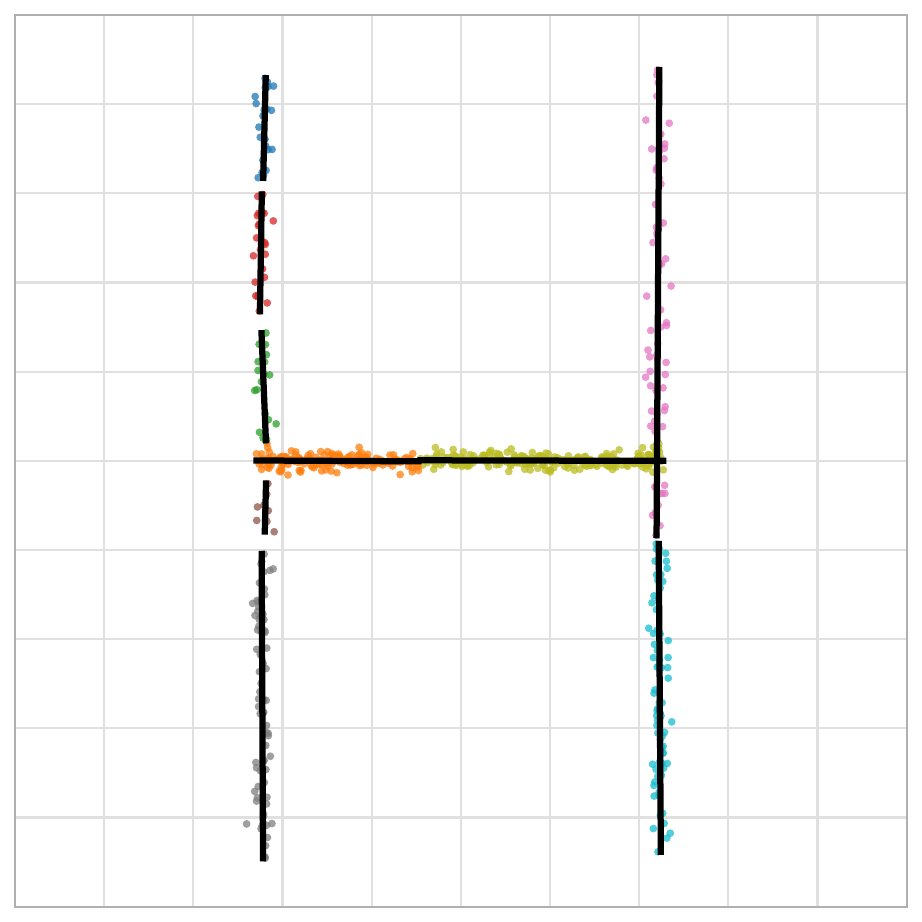}
        \caption{PLSC~\cite{S_and_K21}}
        \label{fig:case3_sk21}
    \end{subfigure}
    \hfill
    \begin{subfigure}[b]{0.235\textwidth}
        \centering
        \includegraphics[width=\textwidth]{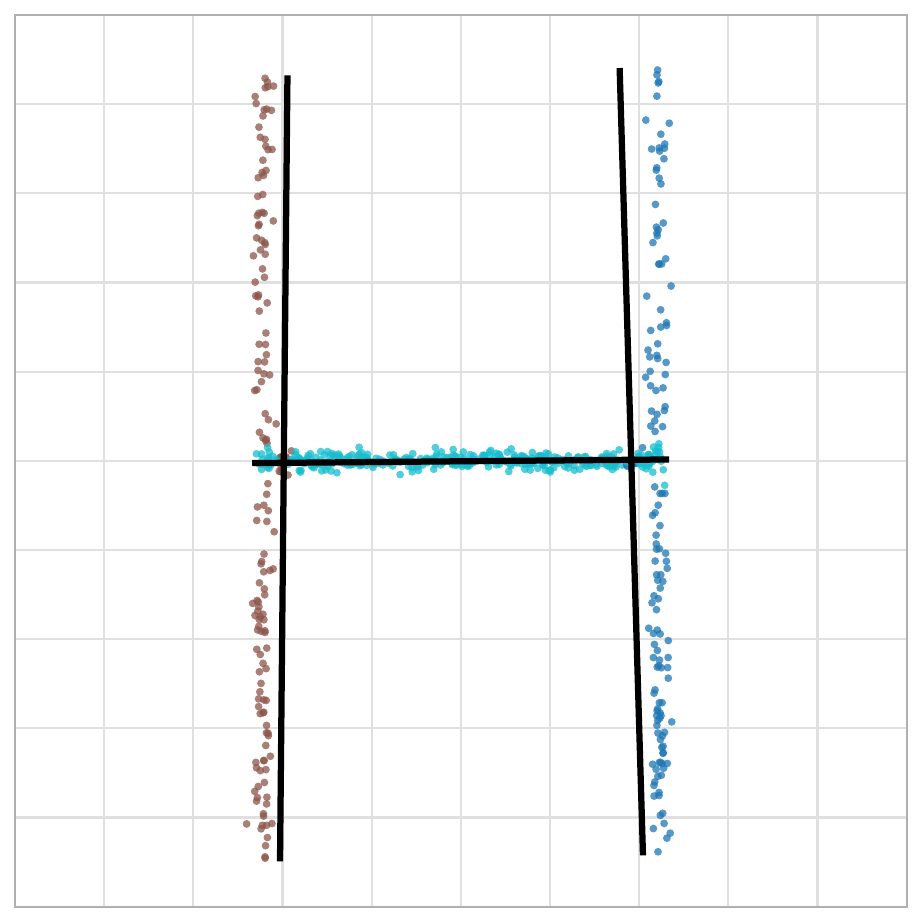}
        \caption{GSFC Pathway 1}
        \label{fig:case3_p1}
    \end{subfigure}
    \hfill
    \begin{subfigure}[b]{0.235\textwidth}
        \centering
        \includegraphics[width=\textwidth]{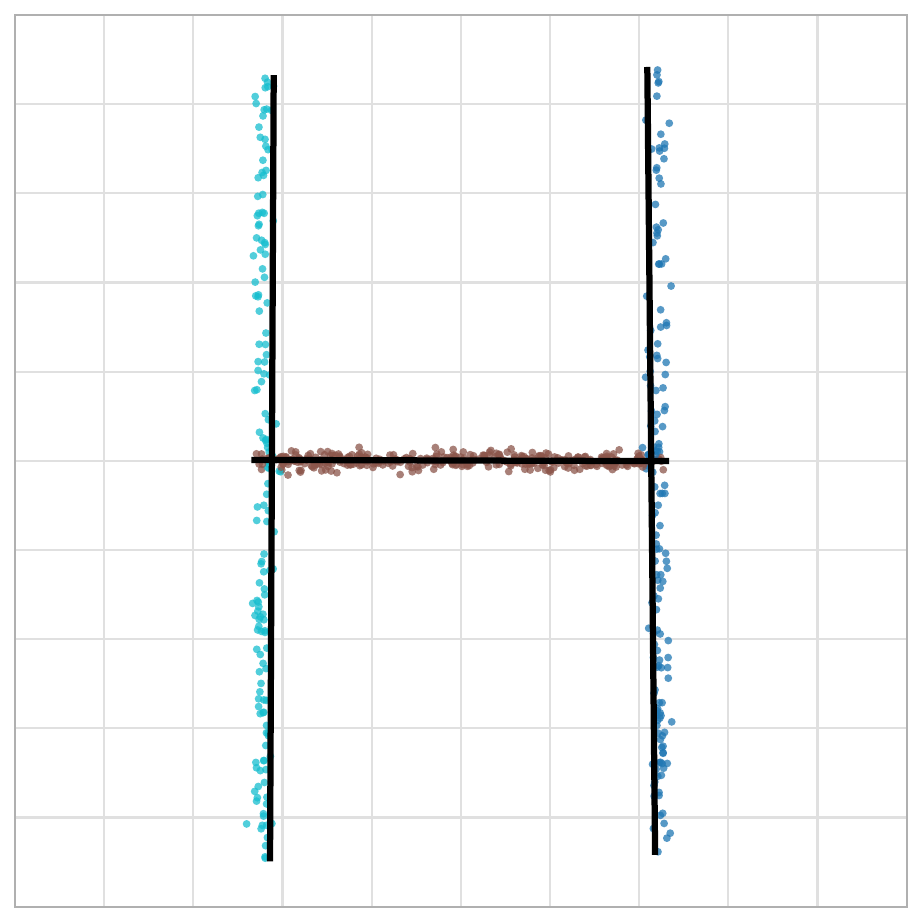}
        \caption{GSFC Pathway 2}
        \label{fig:case3_p2}
    \end{subfigure}
    \hfill
    \caption{Case 5: \textbf{Imbalanced H-shaped Intersection}  The horizontal crossbar contains $2.5 \times$ the data of the vertical legs. The method of \cite{S_and_K21} exhibits severe over-segmentation ($m^*=9$). While both \cite{K18} and the proposed GSFC pathway 1 ($q=1.5$) accurately infer the true cluster cardinality ($m^*=3$), their vertical legs display slight angular distortion induced by the denser crossbar points (albeit pathway 1 enables a slightly tighter reconstruction). In contrast, pathway 2  delivers near-perfect line segment extraction.}
    \label{fig:case3_results}
\end{figure}

Having established the effectiveness of sparsity in the fuzzy clustering pipeline, we now evaluate our framework against two alternative soft sparse clustering approaches for line detection: the possibilistic scheme SPCLS of \cite{K18} and the probabilistic framework  PLSC of \cite{S_and_K21}. To ensure a fair and direct comparison, we adhere as closely as possible to their reference implementations and recommended hyperparameter selections. Rather than evaluating idealized, one-pixel-thin line segments, each segment is modeled as a 2D point cloud with slight spatial noise around the line, simulating imperfect resolution. %We consider diverse geometric configurations featuring acute intersections, junction corners, and imbalanced segment densities, with all competing algorithms over-initialized to $m^{(0)} = 12$ candidate prototypes. The resulting reconstructions are presented in Figs.~3-7, where thick black segments designate the final estimated cluster representatives.
To rigorously assess  performance of GSFC across varied  challenges, we examine 7  images with distinct geometrical properties, initializing each model with $m^{(0)} = 12$ candidate prototypes:

\begin{itemize}
    \item \textbf{Case 1: Three-line Concurrent Intersection (Fig.~\ref{fig:case1_results}).} 
    Three distinct line segments intersect concurrently at the origin with varying angular separations. This setup evaluates whether the models can isolate intersecting linear manifolds without suffering from prototype merging or over-partitioning near the dense center.

    \item \textbf{Case 2: Eight-line Radial Starburst (Fig.~\ref{fig:starburst_comparison}).} This is a more challenging version of the previous experiment.
    Eight symmetric lines intersect at a single central hub, generating an exceptionally dense intersection with acute inter-line angles. This experiment will highlight the importance of sparsity. 
    \item \textbf{Case 3: Overlapping Rectangles / Floppy Disk (Fig.~\ref{fig:floppy_disk}).} 
    This scenario features a closed rectangle inside another rectangle. Sparsity is important, in order to ensure that  nearby inside and outside segments are not conflicted.

    \item \textbf{Case 4: Triangle with Horizontal Intersection (Fig.~\ref{fig:case2_results}).} 
    A closed triangular frame is intersected  by a long horizontal segment, yielding four  intersection points. 

    \item \textbf{Case 5: Imbalanced H-Shaped Configuration with T-Junctions (Fig.~\ref{fig:case3_results}).} 
    Two vertical uprights  (legs) connected by a central horizontal crossbar forming an H. This experiment is particularly interesting because the crossbar   contains $2.5\times$ higher point density than each leg, creating two sharp T-junctions with severe density imbalance. 
    The detector must leverage sparsity to ensure points along the dense interior segment are not erroneously assigned to the legs, where they would act as anchors dragging the uprights into a twisted orientation
\end{itemize}
The results are provided in Figs. 7-11, where thick black segments designate the final estimated cluster representatives.

\noindent \textbf{Remark:} While numerous specialized techniques exist for line segment detection (for example references \cite{LineSeg_Gen1,LineSeg_Gen2,LineSeg_Gen3}) , establishing a state-of-the-art extraction pipeline is beyond the scope of this work. Rather, this application serves as a rigorous geometric testbed to demonstrate that the proposed sparse fuzzy framework structurally outperforms existing possibilistic and probabilistic sparse clustering paradigms.

In summary, visual evaluations across these geometric configurations confirm that the sparse GSFC framework provides a substantial structural advantage over alternative soft-clustering methods. By mathematically enforcing local competition through sparsity, our method successfully resolves complex intersections and bypasses the coincident cluster traps that plague \cite{K18}—a direct consequence of the lack of membership competition in possibilistic frameworks. Furthermore, GSFC prevents the severe over-segmentation seen in the \cite{S_and_K21} probabilistic baseline, whose rigid parametric distribution assumptions restrict its ability to organically adapt to irregular, dense, or non-Gaussian intersecting structures.
\subsubsection{Stress-Testing}
\begin{figure}[!tb]
    \centering
    \includegraphics[width=\linewidth]{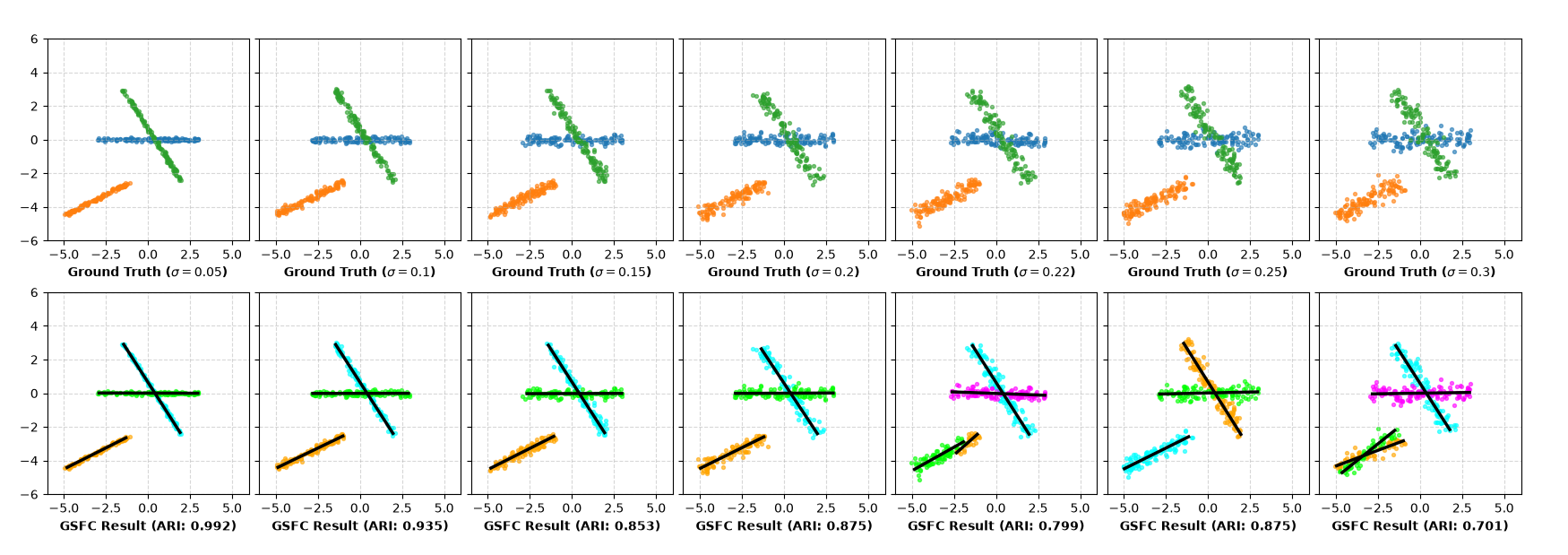}
    \caption{\textbf{Stress-testing GSFC pathway 2 in line detection:} The point generating model of \cite{S_and_K21} is considered with different  segment/cluster variance levels $\sigma$. For variances up to $\sigma=0.2$, the segment reconstruction is near perfect. For $\sigma=0.22$ and $\sigma=0.3 $, there is a small issue in the bottom line segment, which is split in two clusters, but the top intersecting segments are consistently estimated correctly. }
    \label{fig:stress-test-res}
\end{figure}

To evaluate the structural robustness of GSFC under severe spatial noise and understand its limitations, we generated synthetic point clouds comprising three line segments, two of which intersect. Unlike the complex geometric benchmarks evaluated previously (Cases 1-5), this simpler configuration was deliberately selected to isolate the direct impact of point-cloud dispersion ($\sigma$) from compound structural entanglements such as multi-way intersections or asymmetric crossbars. The point distribution strictly follows the generative model proposed in \cite{S_and_K21}, where a 2D dispersion is centered around a line segment parameterized by its endpoints. Stress-testing is conducted by systematically varying the variance $\sigma \in [0.05, 0.30]$ of the points normal to the carrier lines. To rigorously evaluate autonomous cardinality resolution under noisy conditions, GSFC is initialized with an overcomplete set of $m^{(0)} = 10$ candidate prototypes. The results are depicted in Fig. \ref{fig:stress-test-res}, where the robustness is highlighted.  For all noise levels the algorithm maintains sharp, geometrically accurate boundaries at the dense intersection point, correctly identifying the two top segments. For larger values of $\sigma$ ($0.22$ and $0.3$) there is some slight over-segmentation, where the bottom segment is split into two prototypes, potentially requiring a looser merging threshold to account for the high point cloud noise.

\clearpage

\clearpage

\begin{figure*}
    \centering
    \includegraphics[width=\linewidth]{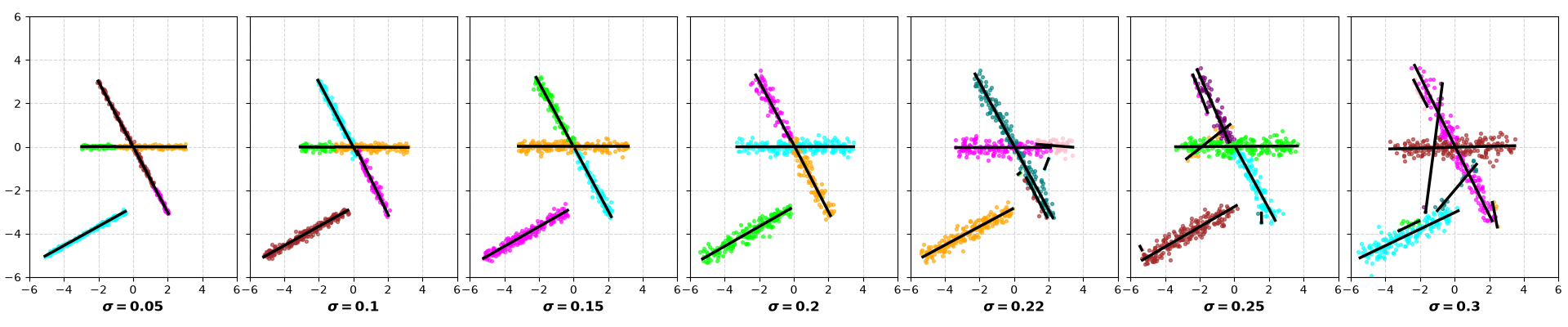} \caption{\textbf{Stress-testing the algorithm of \cite{S_and_K21} for line detection:}  Different  segment/cluster variance levels $\sigma$ are considered for the same point generating model. The algorithm is substantially less robust than GSFC, producing oversegmentation for $\sigma \geq 0.1$, which is severely worsened as $\sigma$ increases.
}\label{fig:stress_test_S_and_K21}
\end{figure*}

\begin{figure*}
    \includegraphics[width=\linewidth]{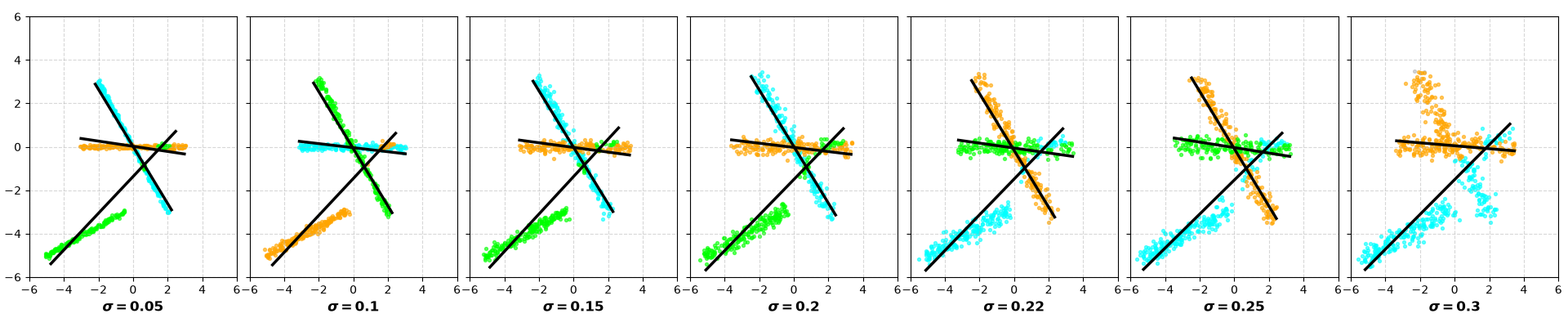}
    \caption{\textbf{Stress-testing the algorithm of \cite{K18} for line detection:}  Different  segment/cluster variance levels $\sigma$ are considered for the same point generating model. While the algorithm can consistently identify the true number of clusters (besides the experiment with the largest $\sigma$), some points are incorrectly assigned to the wrong segment (resulting in incorrectly  estimated and overly long line segments),   with the issue worsening as $\sigma$ increases.
}\label{fig:stress_test_K18}
\end{figure*}
In contrast, in Fig. \ref{fig:stress_test_S_and_K21} we repeated the same experiment for the algorithm proposed in \cite{S_and_K21}. As is evident, for $\sigma>0.1$ the algorithm collapses significantly over-segmenting the line segments and producing too many clusters. Hence, our framework is substantially more robust to noise.

Finally, Fig. \ref{fig:stress_test_K18} illustrates the stress-test evaluation of the algorithm proposed in \cite{K18}. Although the baseline correctly infers the true number of segments for $\sigma < 0.3$, points from the two upper intersecting segments are erroneously assigned to the bottommost cluster. This misclassification severely distorts the estimated orientation of the bottom segment away from its true angle. This phenomenon exacerbates as the variance $\sigma$ increases, progressively degrading the orientation estimates of the intersecting top segments as well. Ultimately, at $\sigma = 0.3$, the algorithm collapses and entirely fails to identify one of the top intersecting segments. %In contrast, GSFC uniquely maintains its structural integrity at this extreme noise level, accurately recovering all segments and yielding a significantly higher ARI.

\subsection{General Discussion}
The empirical evaluation across two synthetic toy datasets,  two tabular real-world dataset, six line-segment detection scenarios, and three real-world HSIs establishes several consistent behavioral characteristics of the proposed GSFC framework. The key takeaways are: 
\begin{itemize}
\item \textbf{Autonomous cardinality recovery:} Across all tested domains the proposed framework consistently and autonomously learns the true number of clusters. By dynamically pruning redundant prototypes, GSFC eliminates the need for prior knowledge of the cluster cardinality, overcoming a major fundamental limitation of baseline methods.
    \item 
\textbf{Sparsity enhances fuzzy fidelity:} Enforcing a sparsity penalty on membership assignments generally yields noticeable performance improvements over standard non-sparsity aware fuzzy clustering. 
\item \textbf{Superiority over standard baselines:} GSFC consistently outperforms traditional hard partitioning (K-means) and GMM, demonstrating stronger resilience to noise, scattered outliers, and structural ambiguity.
\item \textbf{Advancements in elongated cluster case:} The proposed line-detection variant of the framework outperforms existing soft-clustering alternatives in line segment detection. It successfully resolves intersecting geometries, handles dense overlaps, and dynamically prunes redundant prototypes where traditional methods force false alignments.
\item \textbf{Hyperparameter robustness:} The algorithm exhibits   empirical stability regarding the sparsity strength parameter, $s$. Sensitivity analyses confirm that clustering performance remains high and superior to baseline benchmarks across a wide operational range, degrading only when $s$ is pushed to excessively aggressive extremes.
\item \textbf{Robustness to Initialization Conditions:} Interestingly, both GSFC optimization pathways exhibit extreme robustness to initialization.  We hypothesize that initializing with an excess of cluster representatives floods the data space, ensuring every true cluster has representatives placed near or within it from the onset. As dense point clouds drag nearby estimated representatives toward the true cluster centers, coincident representatives are systematically merged, while any isolated representatives stranded away from all clusters are pruned, preventing the algorithm from getting trapped in suboptimal local minima. In contrast, traditional algorithms with a fixed number of clusters can  fail when poorly initialized; for instance, if multiple centroids are inadvertently placed within a single cluster while another cluster receives none, the algorithm can get trapped in a suboptimal partition because it cannot reposition or prune the redundant representatives.
\end{itemize}

We must, however, acknowledge that no learning algorithm is perfect, and our proposed framework certainly exhibits limitations. First, our empirical analysis uncovered a clear boundary condition regarding the choice of the membership exponent: while Pathway~1 ($q=1.5$) generalizes robustly across varying spatial dimensions, Pathway~2 ($q=2.0$) suffers a noticeable performance degradation in higher-dimensional settings (the 30-dimensional WDBC benchmark). As analyzed, quadratic membership decay exacerbates the distance concentration phenomenon, causing artificial cluster representative dispersion away from true positions. Second, because GSFC operates directly on the input feature space in the manner of standard fuzzy and partitioning frameworks, its clustering fidelity remains susceptible to the curse of dimensionality when presented with raw, uncurated data matrices; incorporating automated feature selection or representation learning pipelines constitutes a natural direction to safeguard high-dimensional geometry. Finally, the framework introduces several hyperparameters governing sparsity strength, cluster pruning, and fuzzification. While parameter dependency is a universally acknowledged trait across both fuzzy and sparse clustering paradigms~\cite{K18,S_and_K21,lui2025_fuzzy_with_sparse_noise,Sparse_Fuzzy_with_PCA,Sparce_Fuzzy_SPL,Sparce_Fuzzy_SPL}, our sensitivity analyses confirm that GSFC remains remarkably stable across broad operational ranges of the sparsity parameter $s$ and shows negligible variance across random initialization conditions.
\color{black}
\section{Conclusion and Future Work}
This paper presents the GSFC framework, which effectively addresses the fundamental limitations of dense membership assignments in traditional fuzzy clustering paradigms. By introducing a continuous sparsity penalty—solved via a novel primal-dual bisection methodology—the proposed framework organically drives trace memberships to exactly zero. This  empowers the algorithm to autonomously estimate the true number of clusters by dynamically pruning redundant clusters from an initially overestimated state, bypassing a critical limitation present in nearly all existing fuzzy clustering literature. Furthermore, the integration of a virtual "noise sink" mechanism effectively neutralizes the impact of background outliers. Extensive empirical evaluations demonstrated that GSFC achieves highly competitive performance against traditional benchmarks on both synthetic and real-world datasets. Most notably, when adapted for treating  elongated clusters, the framework successfully outperformed baseline probabilistic and possibilistic alternatives in complex line segment detection scenarios.

Future research will explore integrating the proposed  sparse membership formulation with complementary sparsity paradigms and recent architectural advancements in the clustering literature. Notably, while prior studies predominantly employ sparsity as an external mechanism (such as for sparse noise decomposition, or feature selection), they continue to compute traditional dense fuzzy memberships. A promising direction is to unify these methodologies with GSFC, combining feature or subspace level sparse learning with our sparse partition solver to achieve \textbf{dual-level sparsity} in high-dimensional domains. Furthermore, we envision coupling GSFC with Neural Network (NN)-driven soft clustering frameworks to enhance representation learning across complex manifolds~\cite{DeepClustering1,DeepClustering2}. Finally, we aim to extend the GSFC framework to accommodate large-scale, streaming environments by developing an online optimization variant of our primal-dual bisection algorithm~\cite{ODA_Clustering}.
\printcredits

%% Loading bibliography style file
%\bibliographystyle{model1-num-names}
\bibliographystyle{cas-model2-names}

% Loading bibliography database
\bibliography{cas-refs}
\appendix
\appendix
\section{Computational Complexity Analysis}
\label{app:complexity}

In this section, we analyze the computational complexity of the proposed GSFC framework for both point representatives and the line segment detection adaptation. Let $N$ denote the total number of data points, $d$ the dimensionality of the feature space, and $m^{(0)}$ the initial (overestimated) number of cluster representatives. Because the active cluster count $m$ dynamically decreases via cardinality pruning and coalescence ($m \le m^{(0)}$), the initial allocation $m^{(0)}$ serves as a strict theoretical upper bound on the number of clusters evaluated at any iteration. The parameter $t_{\max}$, defined in Algorithm~\ref{alg:gsfc_main}, denotes the maximum number of iterations allowed for convergence.

\subsection{Point Representatives (Compact Clusters)}
At any iteration of Algorithm~\ref{alg:gsfc_main}, the workload across active clusters is strictly upper-bounded by $m^{(0)}$ across four primary computational stages:
\begin{enumerate}
    \item \textbf{Distance Computation:} Evaluating the regularized Euclidean distances $d(\mathbf{x}_r, \boldsymbol{\theta}_j)$ between all $N$ data points and candidate cluster representatives requires at most $\mathcal{O}(N m^{(0)} d)$ operations.
    \item \textbf{Sparse Membership Evaluation (Algorithm~\ref{alg:membership_update}):} For each sample $\mathbf{x}_r$, the dual variable $\lambda_r$ is determined via an outer bisection search over the interval $[\lambda_{\min}, \lambda_{\max}]$ with stopping tolerance $\epsilon_{bis}$. We introduce $K_{\text{bis}}$ to denote the worst-case number of bisection steps per sample, bounded by:
    \begin{equation}
        K_{\text{bis}} = \left\lceil \log_2\left(\frac{\lambda_{\max} - \lambda_{\min}}{\epsilon_{bis}}\right) \right\rceil.
    \end{equation}
    At each bisection step, evaluating the membership update formulas across all candidate clusters requires at most $\mathcal{O}(m^{(0)})$ operations for the closed-form solutions (Pathways 1 and 2), or $\mathcal{O}(m^{(0)} K_{\text{NR}})$ when utilizing the inner Newton-Raphson solver (where $K_{\text{NR}}$ denotes the average number of Newton iterations). Consequently, updating the membership matrix $\mathbf{U}$ requires at most $\mathcal{O}(N m^{(0)} K_{\text{bis}})$ operations under both analytic pathways.
    \item \textbf{Cluster Update:} Updating the cluster representatives via weighted averages across all $N$ samples incurs a cost of at most $\mathcal{O}(N m^{(0)} d)$.
    \item \textbf{Cardinality Pruning and Merging:} Computing cluster cardinalities requires at most $\mathcal{O}(N m^{(0)})$ operations, while pairwise distance evaluations between active representatives scale as $\mathcal{O}((m^{(0)})^2 d)$. Because $m^{(0)} \ll N$, this step is asymptotically negligible.
\end{enumerate}
Summing the individual contributions, the worst-case overall time complexity of GSFC over $t_{\max}$ iterations is:
\begin{equation}
    \mathcal{O}\Big(t_{\max} N m^{(0)} (d + K_{\text{bis}})\Big).
\end{equation}
Because $t_{\max}$, $m^{(0)}$, and $K_{\text{bis}}$ are fixed hyperparameters, the overall worst-case complexity scales pseudo-linearly with the number of samples $N$. In practice, the effective per-iteration cost progressively decreases as redundant clusters are pruned. ($m \ll m^{(0)}$).
\subsection{Line Segment Adaptation}
When adapting GSFC for line segment detection in 2D edge images ($d = 2$):
\begin{enumerate}
    \item \textbf{Segment Distances:} Computing the perpendicular and endpoint-bounded distances from $N$ edge points to $m$ candidate segments scales as $\mathcal{O}(N m d)$.
    \item \textbf{Membership Evaluation:} Unchanged from the compact formulation, requiring $\mathcal{O}(N m K_{\text{bis}})$ operations.
    \item \textbf{Weighted PCA Updates:} Constructing the $m$ weighted covariance matrices $\mathbf{A}_j$ requires $\mathcal{O}(N m d^2)$ operations, and solving the $d \times d$ eigenvalue problem for each cluster requires $\mathcal{O}(m d^3)$.
    \item \textbf{Pruning and Merging:} Collinear angular and proximity evaluations across candidate segment pairs scale as $\mathcal{O}(m^2 d)$.
\end{enumerate}
For planar image processing where $d = 2$, terms involving $d^2$ and $d^3$ reduce to small constant factors. Thus, the worst-case time complexity of the line segment extraction variant over $t_{\max}$ iterations is:
\begin{equation}
    \mathcal{O}\Big(t_{\max} N m (1 + K_{\text{bis}})\Big) \equiv \mathcal{O}(t_{\max} N m).
\end{equation}

\end{document}